\documentclass{article}

\PassOptionsToPackage{numbers,sort&compress}{natbib}
\usepackage[preprint]{neurips_2026}

\usepackage[utf8]{inputenc}
\usepackage[T1]{fontenc}
\usepackage{url}
\usepackage{booktabs}
\usepackage{amsfonts}
\usepackage{amsmath}
\usepackage{amssymb}
\usepackage{nicefrac}
\usepackage{microtype}
\usepackage{xcolor}
\usepackage{graphicx}
\usepackage{subcaption}
\usepackage{float}
\usepackage{wrapfig}
\usepackage{pifont}

\usepackage{hyperref}
\definecolor{safelink}{HTML}{1F6FB2}
\hypersetup{colorlinks=true, linkcolor=safelink, citecolor=safelink, urlcolor=safelink}
\usepackage[capitalise,nameinlink]{cleveref}

\title{SAFE-Hair: Scalp-Anchored Fields for Exportable Single-View Hair Reconstruction}

\author{%
  Yucheng Wang$^{*}$ \quad Zedong Wang$^{*}$ \quad Yuetong Wu \quad Yue Ma \quad Dan Xu$^{\text{\ding{41}}}$ \\
  The Hong Kong University of Science and Technology \\
  \texttt{\{ywangls, zwangmw, ywufe, ymacn\}@connect.ust.hk} \quad
  \texttt{danxu@cse.ust.hk}
}

\begin{document}

\maketitle
{\renewcommand{\thefootnote}{}\footnotetext{$^{*}$Equal contribution. \quad $^{\text{\ding{41}}}$Corresponding author.}}
\begin{figure*}[h]
    \centering
    \includegraphics[width=\textwidth]{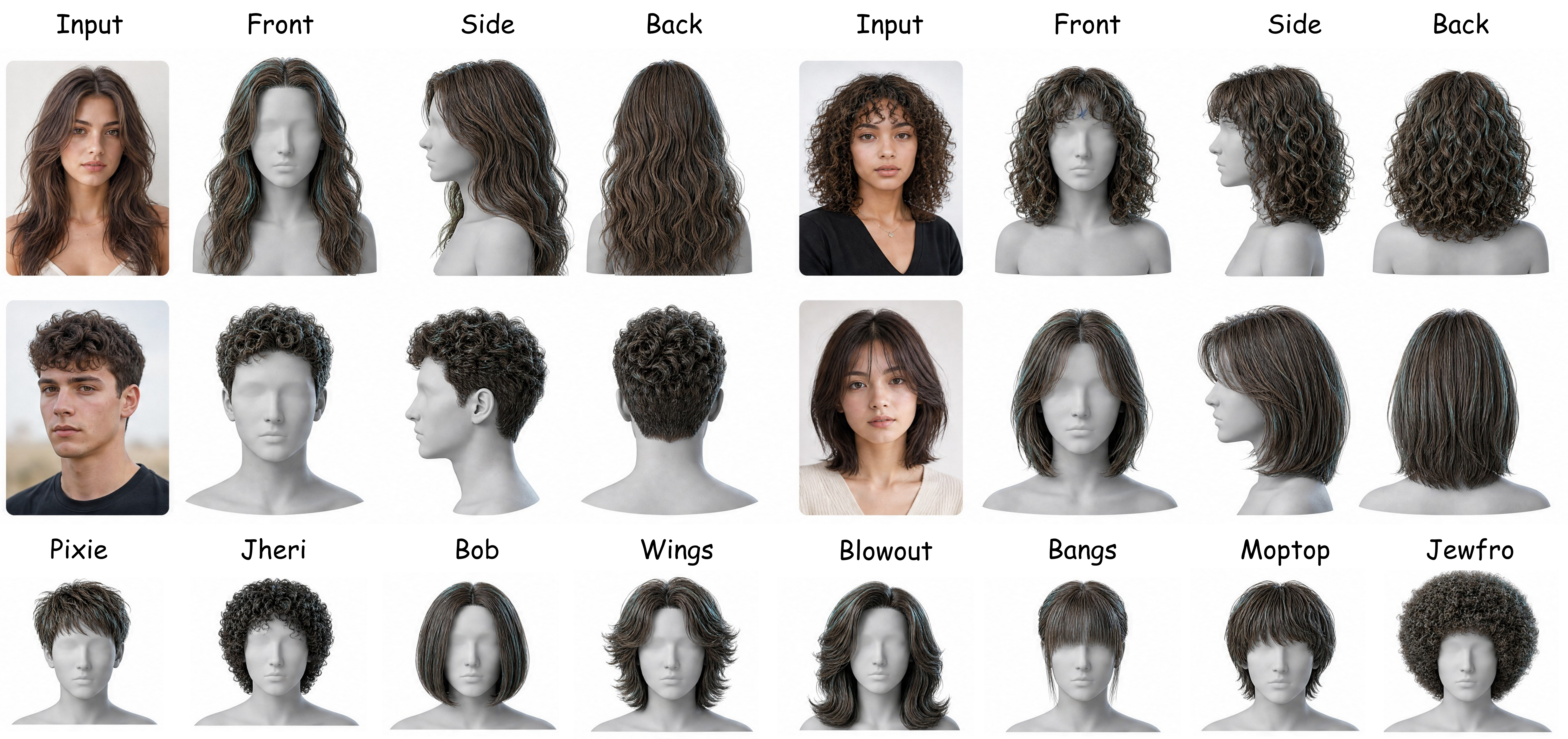}
    \caption{
    From a single portrait image, \textbf{SAFE-Hair} reconstructs an explicit strand-based 3D hair asset with stable scalp-rooted follicles.
We visualize each reconstruction from front, side, and back views, showing complete hairstyle recovery beyond the visible input region.
The bottom row highlights the diversity of generated assets across multiple hairstyle categories.
    }
    \label{fig:teaser}
\end{figure*}

\begin{abstract}
Single-view 3D hair reconstruction has progressed rapidly, but many methods optimize for visual plausibility rather than producing strand assets that can be directly used in grooming, editing, or simulation pipelines. We propose \textbf{SAFE-Hair}, a framework that reconstructs explicit strand grooms from a single portrait by decoding hair from persistent follicles in a canonical scalp-UV domain. Given a fitted template head, SAFE-Hair projects self-supervised image features onto the scalp, uses a conditional Rectified Flow model to generate latent scalp fields, and decodes occupancy, segment directions, length allocation, and total length at a fixed set of scalp follicles. This representation enforces scalp-rooted strand identities by construction and enables direct export with follicle IDs and guide-family metadata. We further introduce attachment, collision, and bending priors to improve geometric validity without per-subject test-time optimization. To evaluate both reconstruction fidelity and asset usability, we introduce \textbf{HairBench-3D}, a benchmark built from public 3D hair assets with standardized alignment, rendering, resampling, and metrics for geometry, direction consistency, invisible-region completion, silhouette agreement, and collision. SAFE-Hair improves Chamfer distance from 3.74 mm to 3.28 mm, F-score from 77.9 to 81.7, invisible-region F-score from 66.1 to 69.8, and penetration ratio from 4.8\% to 3.9\% over the strongest baseline. Project page: \href{https://safe-hair.github.io/}{\textcolor[HTML]{3DA5F5}{safe-hair.github.io}}.

\end{abstract}

\section{Introduction}

Reconstructing 3D hair from a single portrait image is an important step toward scalable digital human creation for visual effects, augmented reality, and telepresence~\citep{wang2025modit}. Recent learning-based methods have made substantial progress in recovering visually plausible hair geometry from monocular inputs. However, visual plausibility alone is often insufficient for downstream content-creation pipelines. In practical grooming and simulation workflows, hair is expected to be represented as strands with well-defined roots on the scalp, stable root identities, and limited interpenetration with the head. In this paper, we distinguish between reconstructing hair geometry and reconstructing a hair asset. A hair asset is not only a set of visually plausible curves or surfaces; it also requires roots that remain attached to a valid scalp domain, persistent strand or follicle identities for editing, and a structure that can be serialized into standard groom pipelines. This distinction is important because many downstream operations—such as guide selection, child-strand interpolation, simulation constraints, collision handling, and artist editing—are defined relative to scalp-rooted strand identities. A reconstruction that lacks these identities may look plausible when rendered but can still be difficult to edit, simulate, or transfer across frames and subjects.

\begin{wrapfigure}{r}{0.48\linewidth}
    \vspace{-2.0em}
    \centering
    \includegraphics[width=\linewidth]{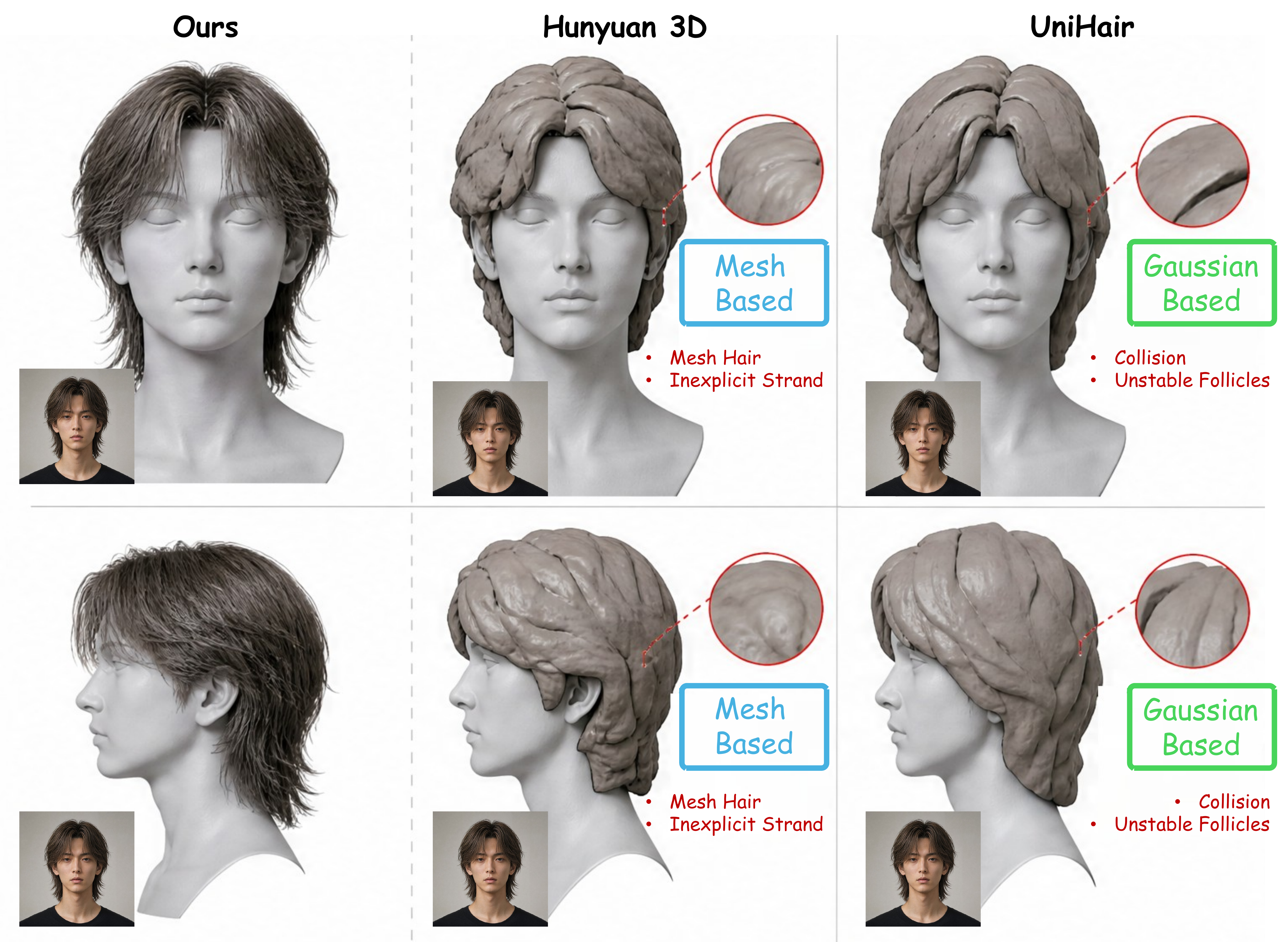}
    \vspace{-1.8em}
    \caption{
    \textbf{Motivation.}
    Existing 3D hair generators often produce rendering-oriented representations: Mesh-based hair~\cite{zhao2025hunyuan3d} lacks explicit strands, while Gaussian-based hair (UniHair~\cite{zheng2024unihair}) lacks persistent follicles and suffers from collisions.
    Our method reconstructs explicit rooted strands with stable scalp follicles, enabling editable and export-ready hair assets from a single portrait.
    }
    \label{fig:motivation}
    \vspace{-2.0em}
\end{wrapfigure}

A second limitation is evaluation. Existing works are commonly evaluated on different datasets, preprocessing pipelines, and metrics. Distance-based or rendering-based metrics are useful, but they do not fully capture whether a reconstructed hairstyle is structurally valid as a strand asset. For example, a method may achieve a low Chamfer distance while placing roots away from the scalp, producing unstable strand identities, or generating strands that penetrate the head. This makes it difficult to compare methods fairly and to measure progress toward exportable hair reconstruction.

To address these limitations, we propose \textbf{SAFE-Hair}, a single-view 3D hair reconstruction framework that produces scalp-indexed strand grooms. Instead of predicting unordered 3D strands directly, SAFE-Hair represents hair in a canonical scalp-UV domain with a fixed set of persistent follicles. Given a portrait image, we fit a template head, warp self-supervised image features into the canonical scalp space, and use a conditional Rectified Flow model~\citep{liu2023rectifiedflow} to generate latent scalp fields. A shared follicle decoder queries these fields at canonical follicles to predict follicle occupancy, strand directions, and explicit length allocation. This scalp-follicle formulation makes root placement valid with respect to the fitted scalp by construction and gives each decoded strand a persistent identity. We further introduce attachment, collision, and bending priors to reduce off-scalp drift, head interpenetration, and implausible local strand deformation. At inference time, SAFE-Hair reconstructs dense strand assets in a single forward pass without test-time optimization. The persistent follicle representation also provides a natural interface for downstream asset construction. Each decoded strand carries a stable follicle index and can be grouped into guide families after reconstruction, allowing the output to be serialized as a structured strand groom. We treat such guide assignments and groom-control attributes as export metadata rather than the primary reconstruction target, and focus our evaluation on geometric fidelity, root consistency, completion, and collision behavior.

We also introduce \textbf{HairBench-3D}, a unified benchmark for evaluating both reconstruction fidelity and asset usability. HairBench-3D is built from public 3D hair resources, including USC-HairSalon~\citep{usc_hairsalon}, Hair20K~\citep{zhou2024hair20k}, and public hair models by Yuksel et al.~\citep{yuksel_hairmodels}. We standardize these assets into a shared head and scalp coordinate system, render consistent single-view inputs, resample strand curves, and evaluate methods using metrics for point-level geometry, local direction consistency, invisible-region completion, silhouette agreement, and head-collision behavior.

Our contributions are summarized as follows:
\begin{enumerate}
    \item We introduce a scalp-anchored representation for single-view 3D hair reconstruction, where strands are decoded from persistent canonical follicles in a shared scalp-UV domain.
    \item We propose SAFE-Hair, a conditional Rectified Flow framework over latent scalp fields coupled with a follicle decoder for predicting strand occupancy, directions, and explicit length allocation without test-time optimization.
    \item We construct exportable strand grooms with stable follicle identities, scalp-attached roots, collision-aware geometry, and guide-family metadata for downstream editing and simulation.
    \item We present HairBench-3D, a unified benchmark with standardized preprocessing and metrics for geometry, direction consistency, invisible-region completion, silhouette agreement, and collision behavior.
\end{enumerate}

\section{Related Work}
\label{sec:related}

\paragraph{3D hair assets and strand representations.}
Explicit strand curves are a natural representation for production hair assets, since they directly encode fiber geometry and support rendering, editing, and simulation~\citep{bergou2008discrete,marschner2003light,chiang2015hair,daviet2023interactive}. Public strand-based datasets, including Yuksel's hair models, USC-HairSalon, and Hair20K, have enabled data-driven reconstruction and generation by providing diverse 3D hairstyles aligned to canonical heads~\citep{usc_hairsalon,yuksel_hairmodels,zhou2024hair20k}. However, learning directly over unordered dense strands remains challenging due to high dimensionality, variable topology, and complex local structures. Recent representations therefore introduce more structured parameterizations, such as guide-and-residual decompositions or differentiable rendering primitives~\citep{he2025perm,zhou2023groomgen,kerbl2023gaussian,zakharov2024gaussianhaircut,luo2024gaussianhair}, in line with structured part-level latents for general 3D asset generation~\citep{dong2025copart}. Our work follows the strand-based direction but represents hairstyles through persistent scalp follicles, producing explicit, editable strands that are directly usable as hair assets.

\paragraph{Single-view hair reconstruction.}
Single-view 3D hair reconstruction is severely ill-posed because much of the hairstyle is self-occluded or ambiguous from one portrait. Early methods address this ambiguity through example-based retrieval, deformation, and user-provided cues~\citep{chai2012singleview,chai2016autohair,hu2015singleview}. Learning-based methods reduce manual interaction by predicting strand features, orientation fields, or intermediate strand/depth maps from image observations~\citep{zhou2018hairnet,zhang2019hairgan,wu2022neuralhdhair,zheng2023hairstep,he2025perm,long2025tangled, jin2026hairorbit}. These methods demonstrate that learned priors can recover plausible monocular hair geometry, but their outputs are often sensitive to image-space cues and may lack stable strand identities, persistent scalp roots, or asset-level validity under occlusion. In contrast, we formulate single-view reconstruction in a canonical scalp-UV domain with fixed follicles, explicitly tying strands to persistent scalp anchors.

\paragraph{Multi-view reconstruction and generative priors.}
Multi-view images or monocular videos provide stronger geometric constraints and have enabled high-fidelity strand reconstruction with learned priors, implicit fields, differentiable rendering, or image-based optimization~\citep{kuang2022deepmvshair,rosu2022neuralstrands,sklyarova2023neuralhaircut,wu2024monohair,takimoto2024drhair,zakharov2024gaussianhaircut,sklyarova2025im2haircut,zhou2024groomcap,luo2012multiview,zhang2017data,nam2019strandaccurate}. Recent generative models further use diffusion or flow-based formulations to learn expressive hairstyle priors for image-conditioned 3D hair generation~\citep{ho2020denoising,rosu2025difflocks,zheng2024unihair,zhou2023groomgen,long2025tangled,lipman2023flowmatching,liu2023rectifiedflow}. Beyond hair, diffusion models also underpin controllable visual generation and editing, including pose-guided and relightable video generation~\citep{ma2024followpose,ma2026livelight,ma2025controllable} and contextual image editing~\citep{wang2026care}. While effective, these approaches often rely on multi-view inputs, per-subject optimization, or rendering. Our goal is complementary: reconstruct production-oriented explicit strand assets from a single portrait, with stable follicle indexing and evaluation criteria that measure asset validity.

\section{Method}
\label{sec:method}

Given a single RGB portrait \(I\), SAFE-Hair reconstructs an explicit strand-based hair asset
\begin{equation}
\mathcal{H} = (\mathcal{S}, \mathcal{R}, \mathcal{E}),
\label{eq:asset_tuple}
\end{equation}
where \(\mathcal{S}\) denotes the reconstructed 3D strands, \(\mathcal{R}\) denotes persistent follicle identities on a canonical scalp, and \(\mathcal{E}\) denotes optional export metadata such as guide-family assignments and groom-control descriptors.
Our objective is to recover a visually plausible hairstyle while producing an asset with valid scalp-rooted strands, stable follicle correspondence, and reduced head interpenetration.
SAFE-Hair follows a scalp-indexed design.
Instead of predicting strands directly in image or world coordinates, we first lift image evidence to a canonical scalp domain, generate dense scalp-aligned latent fields with Rectified Flow~\citep{liu2023rectifiedflow,lipman2023flowmatching}, and then decode explicit strands from a fixed set of persistent follicles.
This design makes root placement valid by construction, keeps strand identities stable across examples, and produces curve-based assets that can be exported without test-time optimization.

\subsection{Overview}
\label{subsec:overview}

\begin{figure}[h]
    \centering
    \includegraphics[width=\linewidth]{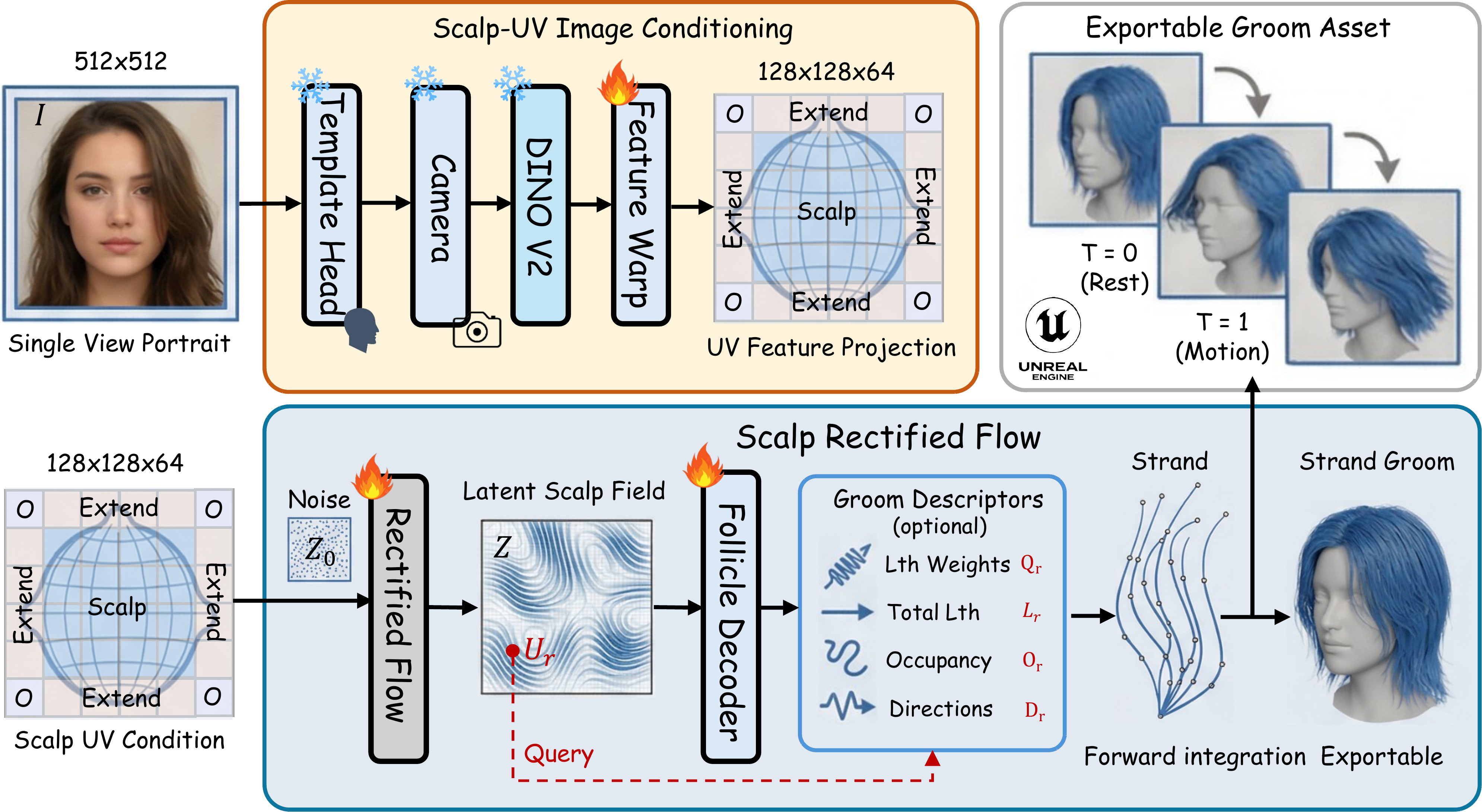}
    \caption{\textbf{Overview of SAFE-Hair.}
    Given a single portrait, we fit a template head and use the recovered camera geometry to project dense image features into a canonical scalp-UV domain, producing a scalp-aligned conditioning map.
A conditional Rectified Flow model transforms Gaussian noise into a latent scalp field \(Z\).
A shared follicle decoder queries \(Z\) at persistent canonical follicles \(\{\mathbf{u}_r\}_{r=1}^{R}\) and predicts occupancy, segment directions, length allocation, total length, and export descriptors.
The final strand curves are obtained by forward integration from fixed scalp roots, yielding explicit scalp-rooted strands with stable follicle identities and groom metadata for downstream export.}
\vspace{-15pt}
    \label{fig:method_flow}
\end{figure}

Shown in Figure~\ref{fig:method_flow}, SAFE-Hair operates in a canonical scalp parameterization shared by all subjects.
Given an input portrait, we first fit a template head and use it as an approximate geometric bridge between the image, the 3D head, and the canonical scalp domain, following common practice in monocular head and hair reconstruction~\citep{sklyarova2023neuralhaircut}.
Sec.~\ref{subsec:representation_conditioning} introduces this representation, including the persistent follicle layout and the projection of dense image features into scalp UV space.
Conditioned on the resulting scalp-aligned feature map, SAFE-Hair generates latent hairstyle fields with the conditional Rectified Flow model described in Sec.~\ref{subsec:flow}.
These fields are decoded into explicit rooted strands by the shared follicle decoder in Sec.~\ref{subsec:decoder}.
Training objectives, including flow matching, strand-level supervision, image-space constraints, and geometric-validity priors, are summarized in Sec.~\ref{subsec:training}.
Finally, Sec.~\ref{subsec:export} describes optional groom-control descriptors and guide-family assignments for export.
Because every active strand remains tied to a fixed follicle identity, the reconstructed asset naturally supports stable correspondence and standard curve-based groom serialization.

\subsection{Canonical scalp space and image conditioning}
\label{subsec:representation_conditioning}

We assume a template head mesh \(\mathcal{M}\) with a scalp region \(\mathcal{M}_{\mathrm{scalp}}\) and a canonical UV parameterization \(\psi : \mathcal{U} \rightarrow \mathcal{M}_{\mathrm{scalp}}\).
After fitting the template to the input portrait with an off-the-shelf monocular head reconstruction method, all hair reasoning is performed in this canonical scalp space rather than directly in the image plane. A key component of our representation is a fixed set of canonical follicles
\begin{equation}
\mathcal{R} = \{\mathbf{u}_r\}_{r=1}^{R}, \qquad \mathbf{x}_{r,1} = \psi(\mathbf{u}_r),
\label{eq:canonical_follicles}
\end{equation}
where \(\mathbf{u}_r\) is a scalp UV location and \(\mathbf{x}_{r,1}\) is its corresponding 3D root position.
The follicle set is shared by all examples, so the index \(r\) serves as a persistent follicle identity.
This choice anchors every decoded strand to a valid scalp location and avoids unconstrained root prediction in 3D space.

To condition the model, we extract dense image features with a self-supervised image encoder and warp them to scalp UV space using the fitted geometry and camera projection.
Self-supervised vision features provide robust dense cues for correspondence and reconstruction tasks~\citep{oquab2024dinov2}.
The resulting scalp-aligned conditioning map \(C\) places visible image evidence in a canonical support, where it can be fused with learned hairstyle priors to complete ambiguous or invisible regions.

\subsection{Rectified-flow generation of scalp fields}
\label{subsec:flow}

For each training hairstyle, we first convert its strand geometry into a canonical scalp-aligned supervision field.
Each ground-truth strand is assigned to its nearest canonical follicle according to geodesic distance on the scalp.
For an occupied follicle, the associated strand is resampled to a fixed number of control points and encoded with its occupancy, local segment directions, total length, and normalized segment-length allocation.
These per-follicle quantities are rasterized to a UV grid and compressed into a latent scalp field \(z_1 \in \mathbb{R}^{H \times W \times C_z}\).
Empty follicles contribute occupancy labels but no geometric targets.
Further details of the canonical field construction are provided in Appendix~\ref{app:field_construction}.

We model the distribution of latent scalp fields with Rectified Flow~\citep{liu2023rectifiedflow,lipman2023flowmatching}.
Given the scalp-aligned conditioning map \(C\), a Gaussian sample \(z_0 \sim \mathcal{N}(0,I)\), a training target \(z_1\), and \(t \sim \mathcal{U}(0,1)\), Rectified Flow constructs the interpolation
\begin{equation}
    z_t = (1-t)z_0 + t z_1 .
\end{equation}
We train a conditional vector field \(v_{\theta}\) to predict the transport direction from \(z_0\) to \(z_1\):
\begin{equation}
    \mathcal{L}_{\mathrm{rf}}
    =
    \mathbb{E}_{z_0,z_1,t}
    \left[
    \left\|
    v_{\theta}(z_t,t,C) - (z_1-z_0)
    \right\|_2^2
    \right].
\end{equation}
At inference time, we sample \(z_0 \sim \mathcal{N}(0,I)\) and integrate
\begin{equation}
    \frac{d z_t}{dt} = v_{\theta}(z_t,t,C), \qquad Z=z_{t=1},
\end{equation}
using a fixed-step ODE solver.
Unless otherwise specified, a single sample is used for all evaluations.

\subsection{Persistent follicle decoding}
\label{subsec:decoder}

Given the generated latent scalp field \(Z\), we decode one candidate strand at each canonical follicle.
For follicle \(r\), we extract a local feature
\begin{equation}
\mathbf{h}_r = Q(Z, \mathbf{u}_r),
\label{eq:follicle_feature}
\end{equation}
where \(Q\) denotes bilinear sampling with local aggregation in scalp UV space.
A shared decoder predicts follicle occupancy, strand geometry, and optional groom-control descriptors:
\begin{equation}
    (o_r, \mathbf{q}_r, \mathbf{D}_r, L_r, \mathbf{e}_r) = D_{\phi}(\mathbf{h}_r),
\end{equation}
where \(o_r \in [0,1]\) is the occupancy probability,
\(\mathbf{q}_r\) are normalized segment-length weights,
\(\mathbf{D}_r=\{\hat{\mathbf{d}}_{r,t}\}_{t=1}^{T-1}\) are unit segment directions,
\(L_r\) is the total strand length, and \(\mathbf{e}_r\) is an optional descriptor used only for export metadata.
We enforce \(\mathbf{q}_r\) with a softmax, normalize predicted directions to unit length, and constrain \(L_r\) to be positive with a softplus output.

The strand control points are obtained by forward integration from the scalp root:
\begin{equation}
    \ell_{r,t}=L_r q_{r,t}, \qquad
    \mathbf{x}_{r,t+1}
    =
    \mathbf{x}_{r,t}
    +
    \ell_{r,t}\hat{\mathbf{d}}_{r,t}.
\end{equation}
A strand is kept in the final set if \(o_r\) exceeds a threshold.
This factorization separates root position, total length, and local segment allocation, which makes the reconstructed strand explicitly scalp-rooted and prevents length from being represented only by unconstrained coordinate offsets.

\subsection{Training objectives and geometric priors}
\label{subsec:training}

SAFE-Hair is trained with a combination of generative, strand-level, image-space, and geometric-validity objectives.
For 3D hair assets, we supervise follicle occupancy, strand directions, length allocation, total length, and reconstructed control points in canonical follicle space.
For real portraits, we render the reconstructed strands and compare them to image using silhouette and orientation cues:
\begin{equation}
\mathcal{L}_{\mathrm{img}}
=
\lambda_{\mathrm{sil}} \mathcal{L}_{\mathrm{sil}}
+
\lambda_{\mathrm{ori}} \mathcal{L}_{\mathrm{ori}}.
\label{eq:image_loss}
\end{equation}

To improve asset validity, we regularize decoded strands with lightweight attachment, collision, and bending priors.
The attachment prior encourages the first segment of a strand to emerge outward from the scalp surface.
The collision prior penalizes strand samples that intersect or approach the fitted head geometry too closely.
The bending prior discourages abrupt local kinks while still allowing naturally curved hairstyles.
These priors reduce root flipping, head interpenetration, and unstable strand shapes without requiring test-time optimization. Optional groom-control descriptors are supervised when corresponding geometry-derived targets or metadata are available.
In our implementation, such descriptors may include rest length, curl, or other low-dimensional quantities derived from the reconstructed curve.
We treat them as export metadata rather than as the primary target.

The full training objective is
\begin{equation}
\mathcal{L}
=
\lambda_{\mathrm{rf}}\mathcal{L}_{\mathrm{rf}}
+
\lambda_{\mathrm{strand}}\mathcal{L}_{\mathrm{strand}}
+
\lambda_{\mathrm{img}}\mathcal{L}_{\mathrm{img}}
+
\lambda_{\mathrm{valid}}\mathcal{L}_{\mathrm{valid}}
+
\lambda_{\mathrm{desc}}\mathcal{L}_{\mathrm{desc}}.
\label{eq:total_loss}
\end{equation}
Here \(\mathcal{L}_{\mathrm{strand}}\) includes occupancy, length, direction, and point-level supervision, while \(\mathcal{L}_{\mathrm{valid}}\) groups the attachment, collision, and bending priors.
In practice, synthetic or 3D-asset samples provide direct strand supervision, while real portraits mainly contribute image-space supervision together with the shared geometric-validity priors.

\subsection{Exportable groom construction}
\label{subsec:export}

Because each strand is decoded from a canonical follicle \(\mathbf{u}_r\), its root identity is the follicle index \(r\), which remains stable across examples and downstream processing.
For each active strand, we export its control points, follicle ID, scalp coordinate, and optional groom descriptor \(\mathbf{e}_r\), providing a stable interface for editing, correspondence, and serialization.

We optionally construct guide families by selecting guide strands from active follicles and assigning each remaining strand to the nearest guide in a joint scalp-position and strand-feature space:
\begin{equation}
g(r) = \arg\min_{m \in \{1,\dots,G\}}
\left[
\alpha \, \| \mathbf{u}_r - \mathbf{u}_m \|_2^2
+
\beta \, \| \mathbf{f}_r - \mathbf{f}_m \|_2^2
\right],
\label{eq:guide_assignment}
\end{equation}
where \(\mathbf{f}_r\) is a low-dimensional strand descriptor.
The assignment is used only as export metadata and does not affect the reconstructed geometry. The final curves can be serialized into standard groom formats such as Alembic or engine-native strand assets. At inference time, SAFE-Hair produces dense strand assets in a single forward pass without test-time optimization.

\section{Experiments}
\label{sec:experiments}

\subsection{HairBench-3D and Evaluation Protocol}
\label{subsec:benchmark_protocol}

We evaluate on HairBench-3D, a benchmark for single-view 3D hair reconstruction.
HairBench-3D compares methods with different output representations, including strands, meshes, implicit fields, and other 3D hair proxies, under a unified evaluation protocol with identical inputs, camera parameters, head alignment, and metric implementations.
Benchmark construction, asset filtering, registration, visibility labeling, metric definitions, and output conversion are detailed in Appendix~\ref{app:hairbench}.

\paragraph{Held-out test split.}
\begin{wraptable}{r}{0.48\linewidth}
\vspace{-0.8em}
\centering
\small
\caption{\textbf{HairBench-3D held-out test split.}
The split is defined at the hairstyle-asset level.}
\label{tab:hairbench_split}
\begin{tabular}{lcl}
\toprule
Source & \#Assets & Main role \\
\midrule
Hair20K~\cite{zhou2024hair20k} & 1,200 & Diverse hairstyles \\
USC-HairSalon~\cite{usc_hairsalon} & 343 & Groom assets \\
Yuksel~\cite{yuksel_hairmodels} & 8 & Varied examples \\
\midrule
Total & 1,551 & Frozen test set \\
\bottomrule
\end{tabular}
\vspace{-1.0em}
\end{wraptable}
HairBench-3D contains 1,551 frozen test hairstyles collected from multiple strand-level sources, including Hair20K, USC-HairSalon, and Yuksel Hair Meshes~\citep{zhou2024hair20k, usc_hairsalon, yuksel_hairmodels}. 
The test split is fixed before model development and used only for final evaluation.
No hairstyle asset, rendered view, augmented variant, or near-duplicate from the test split is used for training, validation, hyperparameter tuning, data filtering, or qualitative example selection.
We describe the asset-level separation and near-duplicate removal procedure in Appendix~\ref{app:hairbench_split}. We will release the whole \textbf{HairBench-3D} together with the metrics script to HuggingFace soon.

\paragraph{Canonical evaluation.}
All assets are registered to a shared template head and scalp-UV parameterization.
Each valid strand root is projected to the scalp surface and assigned to a canonical follicle.
All methods are evaluated in the same canonical 3D coordinate system, with distance-based metrics reported in millimeters.
For each test asset, we render the same calibrated near-frontal RGB input, hair mask, and orientation map for all methods.
Held-out side and back views are used only for render-space metrics and are never provided during inference.

\begin{figure}[t]
    \centering
    \includegraphics[width=\linewidth]{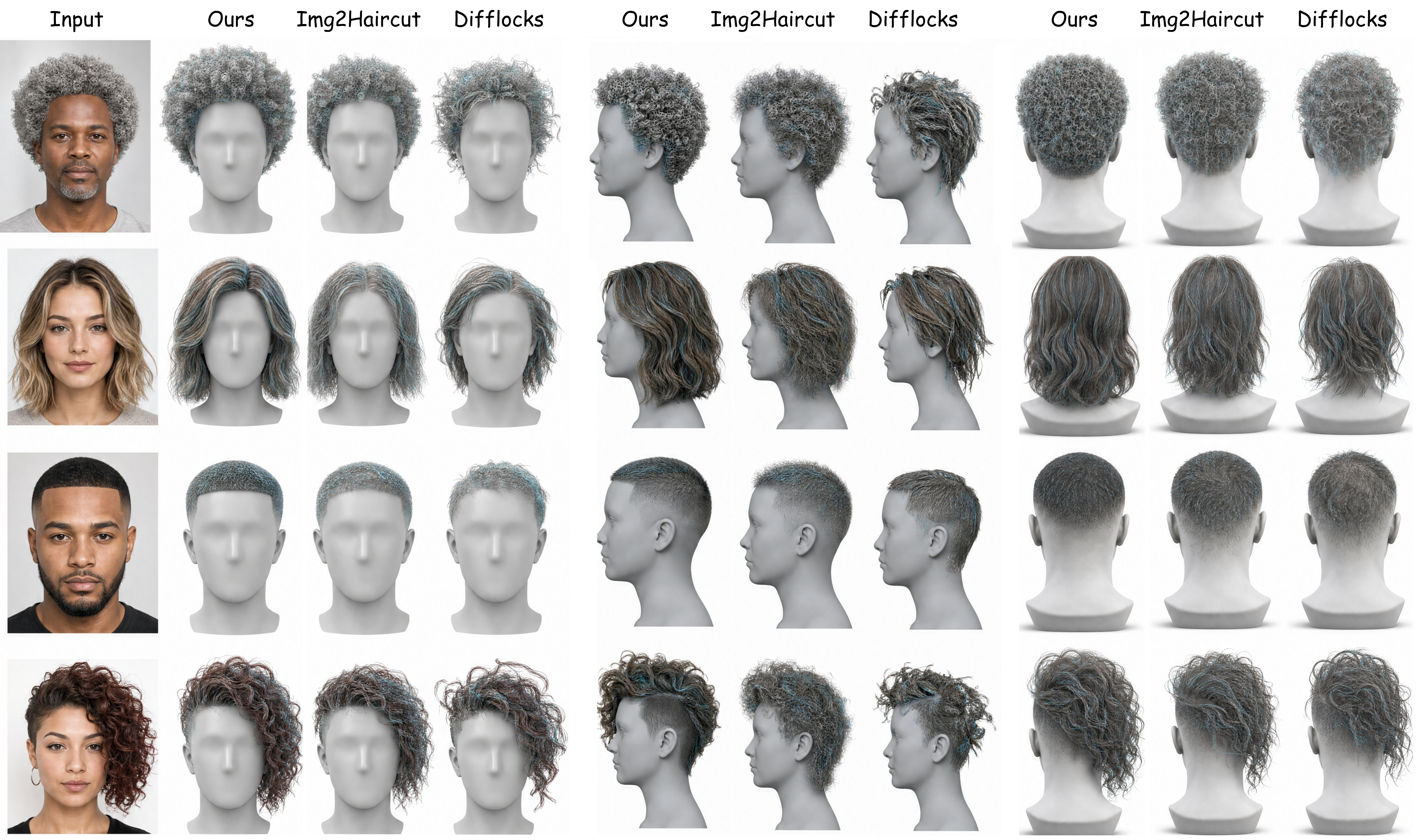}
    \caption{\textbf{Qualitative comparison.}
    For each input image, we show reconstructed hair from multiple viewpoints.
    SAFE-Hair produces scalp-attached strand geometry with more consistent back-view completion and fewer head-interpenetration artifacts than prior methods.}
    \label{fig:Main}
\end{figure}

\paragraph{Metrics and fairness.}
We report five primary metrics: ChamferPts, F-score@4mm/40$^\circ$, Invisible-region F-score, Silhouette IoU, and Penetration Ratio.
These metrics measure point-level geometry, tangent-aware reconstruction quality, invisible-region completion, rendered silhouette agreement, and physical validity, respectively.
For baselines, we use official pretrained weights and inference code when available.
Since prior methods rely on different private, synthetic, or partially released training collections, we enforce fairness through a fixed held-out test split, identical inputs, identical canonical alignment, and representation-independent metric conversion.
Formal metric definitions and baseline conversion details are provided in Appendix~\ref{app:metrics} and Appendix~\ref{app:baseline_protocol}.

\subsection{Implementation Details}
\label{subsec:implementation}

\begin{table}[t]
    \centering
    \small
    \resizebox{\textwidth}{!}{
    \begin{tabular}{l c c c c c}
        \toprule
        Method 
        & ChamferPts $\downarrow$ 
        & F-score $\uparrow$ 
        & F-score (Invisible) $\uparrow$ 
        & IoU $\uparrow$ 
        & Penetration Ratio $\downarrow$ \\
        \toprule
        Perm~\citep{he2025perm} 
        & 5.82 & 61.4 & 42.7 & 0.781 & 8.6 \\
        UniHair~\citep{zheng2024unihair} 
        & 4.85 & 69.2 & 57.0 & 0.807 & 26.8 \\
        NeuralHDHair~\citep{wu2022neuralhdhair} 
        & 4.36 & 71.8 & 56.2 & 0.824 & 6.1 \\
        Hairstep~\citep{zheng2023hairstep} 
        & 4.48 & 70.6 & 59.4 & 0.872 & 7.2 \\
        DiffLocks~\citep{rosu2025difflocks} 
        & 3.91 & 76.5 & 63.8 & 0.847 & 5.4 \\
        Im2Haircut~\citep{sklyarova2025im2haircut} 
        & 3.74 & 77.9 & 66.1 & 0.861 & 4.8 \\
        \toprule
        Ours 
        & \textbf{3.28} 
        & \textbf{81.7} 
        & \textbf{69.8} 
        & \textbf{0.884} 
        & \textbf{3.9} \\
        Ours (5 seeds)
        & 3.29\,$\pm$\,0.03 & 81.6\,$\pm$\,0.2 & 69.7\,$\pm$\,0.3 & 0.883\,$\pm$\,0.001 & 4.0\,$\pm$\,0.15 \\
        \toprule
    \end{tabular}
    }
    \caption{\textbf{Main results on HairBench-3D.}
    Lower is better for ChamferPts and Penetration Ratio, while higher is better for F-score@4mm/40$^\circ$, Invisible-region F-score, and Silhouette IoU.
    ``Ours'' uses one fixed seed; ``Ours (5 seeds)'' reports mean $\pm$ standard deviation over five Rectified Flow sampling seeds with fixed weights.}
    \vspace{-20pt}
    \label{tab:hairbench_main}
\end{table}

\paragraph{Training data.}
SAFE-Hair is trained on an independent dataset that contains large-scale synthetic strand renderings and curated high-quality hairstyles.
No assets from HairBench-3D are used for training, validation, or hyperparameter selection.
Following recent strand-based reconstruction methods~\cite{sklyarova2025im2haircut, he2025perm}, our model learns hair priors from synthetic and curated hairstyles.
After filtering invalid roots, extremely short strands, and assets with severe scalp misalignment, we split the independent dataset into 18,000 training hairstyles and 1,000 validation hairstyles.
HairBench-3D is used exclusively for evaluation, and all benchmark results are reported on its 1,551-style test split. 

\paragraph{Model and optimization.}
We use \(N=32{,}768\) canonical follicles and resample each active strand to \(T=32\) control points.
Input images are cropped to \(512\times512\), the scalp feature map has resolution \(128\times128\), and the latent scalp field has \(C=64\) channels.
We train for 300k iterations using AdamW with learning rate \(2\times10^{-4}\), weight decay \(10^{-4}\), batch size 8, and a cosine learning-rate schedule with 10k warm-up iterations.
Training is performed on 4 NVIDIA L40 GPUs and takes around 45 hours.
Unless otherwise stated, we use
\(\lambda_{\mathrm{rf}}=0.1\),
\(\lambda_{\mathrm{strand}}=1.0\),
\(\lambda_{\mathrm{img}}=0.1\),
\(\lambda_{\mathrm{valid}}=0.05\), and
\(\lambda_{\mathrm{desc}}=0.01\).
Weights are selected on the validation split and fixed for all experiments.

\paragraph{Inference.}
At inference time, SAFE-Hair reconstructs dense scalp-anchored strands in a single forward pass without test-time optimization.
All inference results are generated on a single NVIDIA RTX 3090 GPU.
For a \(512\times512\) input image with \(32{,}768\) follicles, \(T=32\) control points per strand, and 20 Euler steps, inference takes 4.2 seconds on average, excluding optional file export.
The effect of the number of sampling steps and of the output resolution is analyzed in Appendix~\ref{app:steps_scalability}.

\subsection{Baselines}
\label{subsec:baselines}

We compare against representative single-view hair reconstruction methods spanning several output representations and inference paradigms, including regression-based strand prediction, implicit reconstruction, hair-growing methods, diffusion-based generation, and optimization-based fitting.
Our main comparisons include Perm~\citep{he2025perm}, UniHair~\citep{zheng2024unihair}, NeuralHDHair~\citep{wu2022neuralhdhair}, Hairstep~\citep{zheng2023hairstep}, DiffLocks~\citep{rosu2025difflocks}, and Im2Haircut~\citep{sklyarova2025im2haircut}.
For methods whose native outputs are not dense strands, we convert their results to the common benchmark representation required by each metric.
For mesh-based outputs, we uniformly sample surface points and estimate local directions when applicable.
For strand-based outputs, we resample strands to the benchmark control-point convention before computing metrics.

\begin{figure}[t]
    \centering
    \includegraphics[width=\textwidth]{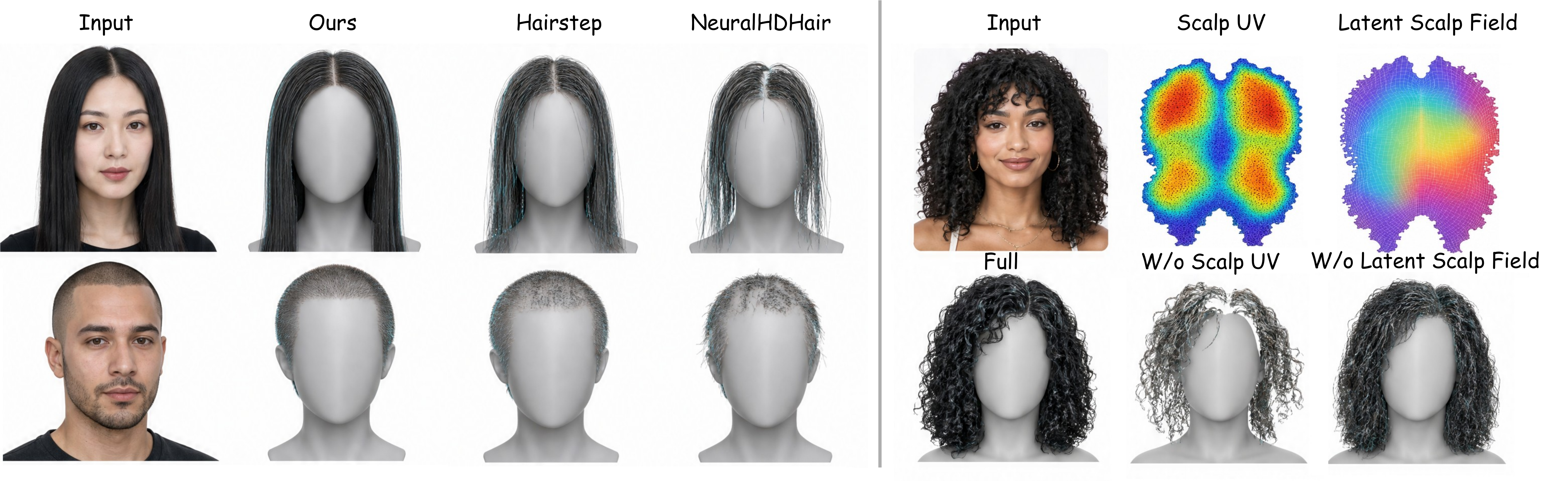}
    \caption{
    \textbf{Qualitative ablation.}
    \textbf{Left:} SAFE-Hair produces more complete and coherent scalp-rooted strands.
    \textbf{Right:} removing scalp-UV conditioning or the latent scalp field degrades strand coherence and hairstyle consistency.}
    \label{fig:qual_ablation}
\end{figure}

\begin{table}[t]
    \centering
    \small
    \resizebox{\textwidth}{!}{
    \begin{tabular}{l c c c c c}
        \toprule
        Variant 
        & ChamferPts $\downarrow$ 
        & F-score $\uparrow$ 
        & F-score (Invisible) $\uparrow$
        & IoU $\uparrow$ 
        & Penetration Ratio $\downarrow$ \\
        \midrule
        W/o Scalp UV 
        & 3.91 & 77.8 & 63.5 & 0.857 & 5.1 \\
        W/o Rectified Flow 
        & 3.67 & 78.9 & 64.1 & 0.866 & 4.8 \\
        W/o Latent Scalp Field 
        & 3.58 & 79.6 & 66.0 & 0.871 & 4.6 \\
        W/o Length Factorization 
        & 3.49 & 80.2 & 67.1 & 0.876 & 4.4 \\
        W/o Directions 
        & 3.43 & 80.8 & 68.2 & 0.879 & 6.7 \\
        \midrule
        Full SAFE-Hair 
        & \textbf{3.28} 
        & \textbf{81.7} 
        & \textbf{69.8}
        & \textbf{0.884} 
        & \textbf{3.9} \\
        \bottomrule
    \end{tabular}
    }
    \vspace{5pt}
    \caption{\textbf{Ablation study on HairBench-3D.}
    We evaluate the contribution of the key design choices in SAFE-Hair using the primary benchmark metrics, including geometry accuracy, visible and invisible-region reconstruction quality, silhouette consistency, and physical validity.}
    \label{tab:ablation_main}
\end{table}

\subsection{Main Results}
\label{subsec:main_results}

Table~\ref{tab:hairbench_main} shows that SAFE-Hair performs best across all five primary metrics.
Compared with the strongest prior baseline, SAFE-Hair reduces ChamferPts from 3.74 to 3.28 and improves F-score@4mm/40$^\circ$ from 77.9 to 81.7, indicating more accurate recovery of both geometry and local strand direction.
The gain in Invisible-region F-score suggests stronger completion in occluded and back-view regions from a single input image.
SAFE-Hair also achieves the highest Silhouette IoU, showing better recovery of the global hairstyle outline, and the lowest Penetration Ratio, indicating fewer interpenetration artifacts with the head or body proxy. 

\begin{table}[t]
    \centering
    \small
    \begin{tabular*}{\textwidth}{@{\extracolsep{\fill}} l c c c c}
        \toprule
        Method 
        & RootDist $\downarrow$ 
        & ValidRoot $\uparrow$ 
        & FollicleStab. $\uparrow$ 
        & ExportSucc. $\uparrow$ \\
        \midrule
        Perm~\citep{he2025perm} 
        & -- & -- & -- & -- \\
        NeuralHDHair~\citep{wu2022neuralhdhair} 
        & 3.8 & 82.1 & -- & 71.3 \\
        Hairstep~\citep{zheng2023hairstep} 
        & 2.9 & 87.4 & -- & 76.8 \\
        DiffLocks~\citep{rosu2025difflocks} 
        & 2.4 & 89.6 & 0.61$^\dagger$ & 82.5 \\
        Im2Haircut~\citep{sklyarova2025im2haircut} 
        & 1.7 & 93.2 & 0.68$^\dagger$ & 88.1 \\
        \midrule
        Ours 
        & \textbf{0.2} 
        & \textbf{99.8} 
        & \textbf{0.91} 
        & \textbf{99.6} \\
        \bottomrule
    \end{tabular*}
    \vspace{5pt}
    \caption{\textbf{Asset usability on HairBench-3D.}
    RootDist measures average root-to-scalp distance in mm; ValidRoot and ExportSucc.\ are percentages.
    ``--'' marks metrics that are not applicable, and $^\dagger$ indicates approximate follicle stability computed by nearest-root matching.
    Metric definitions are given in Appendix~\ref{app:asset_usability_metrics}.}
    \label{tab:asset_usability}
\end{table}

\begin{table}[t]
    \centering
    \small
    \begin{tabular*}{\textwidth}{@{\extracolsep{\fill}} l c c c c}
        \toprule
        Input yaw 
        & ChamferPts $\downarrow$ 
        & F-score $\uparrow$ 
        & F-score (Invisible) $\uparrow$ 
        & IoU $\uparrow$ \\
        \midrule
        $0^\circ$ 
        & 3.28 & 81.7 & 69.8 & 0.884 \\
        $30^\circ$ 
        & 3.35 & 81.0 & 68.9 & 0.878 \\
        $60^\circ$ 
        & 3.57 & 79.4 & 66.2 & 0.861 \\
        $90^\circ$ 
        & 3.86 & 77.2 & 62.8 & 0.839 \\
        \bottomrule
    \end{tabular*}
    \vspace{5pt}
    \caption{\textbf{Generalization to side-view inputs on HairBench-3D.}
    Each held-out hairstyle is rendered at four input yaw angles and reconstructed from the corresponding portrait.}
    \label{tab:yaw}
\end{table}

In addition to geometric fidelity, we evaluate whether the reconstructed hairstyle behaves as a scalp-rooted strand asset. Table~\ref{tab:asset_usability} reports root-to-scalp distance, valid-root ratio, follicle stability under small input perturbations, and export success rate. SAFE-Hair achieves high root validity because roots are decoded from fixed canonical follicles rather than predicted as unconstrained 3D points. The active follicle set is also more stable under crop, color, and mask perturbations, which is important for editing and potential temporal use. Table~\ref{tab:yaw} reports side-view inputs: performance degrades gradually with yaw, and at \(60^\circ\) it still matches the strongest baseline's frontal-view results. These results support the central claim that SAFE-Hair reconstructs not only plausible hair geometry but also an exportable groom structure.

Figures~\ref{fig:Main} and~\ref{fig:qual_ablation} show representative reconstructions exported to Unreal Engine.
From a single portrait, SAFE-Hair produces view-consistent strand geometry with plausible back-of-head completion and stable scalp attachment.
Compared with prior approaches that often oversmooth occluded regions or produce loose root placement, our results better preserve strand structure, silhouette fidelity, and exportable scalp-rooted geometry. Additional results are shown in Appendix~\ref{app:additional_qualitative}.

\subsection{Ablation Studies}
\label{subsec:ablation}

Table~\ref{tab:ablation_main} analyzes the contribution of each component.
Removing scalp-UV conditioning causes the largest degradation, increasing ChamferPts from 3.28 to 3.91 and reducing Invisible-region F-score from 69.8 to 63.5.
This shows that the canonical scalp-UV domain is crucial for stable root localization and coherent completion, especially in invisible side and back regions.
Without this anchoring, predicted strands become more weakly attached and less spatially organized.

Removing rectified flow also degrades geometry and invisible-region completion, indicating that flow-based distributional refinement helps infer plausible structures beyond the visible input.
The latent scalp field improves geometry and silhouette consistency by sharing spatially coherent features across neighboring scalp regions.
As shown in Figure~\ref{fig:qual_ablation}, removing the latent field preserves the coarse hairstyle shape but weakens local strand coherence and volume consistency.

Without length factorization, performance drops for hairstyles with large strand-length variation, leading to worse ChamferPts and Invisible-region F-score.
Removing directional supervision has a smaller effect on point-level geometry but increases Penetration Ratio from 3.9 to 6.7, suggesting that tangent-aware prediction is important for physically plausible growth directions near the scalp.

\section{Conclusion and Limitations}
\label{sec:conclusion}

We introduced SAFE-Hair, a single-view hair reconstruction framework for export-ready strand assets. By reconstructing hair in a canonical scalp-UV domain, generating latent scalp fields with Rectified Flow, and decoding strands from persistent canonical follicles, SAFE-Hair produces rooted strands with stable follicle identities, guide--child organization, and simulation-oriented attributes. Attachment, collision, and bending priors further improve structural validity by reducing off-scalp drift and head interpenetration. We also proposed HairBench-3D, a unified benchmark for evaluating single-view 3D hair reconstruction across strands, meshes, implicit fields, and other hair proxies. With standardized preprocessing, rendering, alignment, and metrics, HairBench-3D enables controlled comparison under identical inputs and evaluation code. Experiments demonstrate that SAFE-Hair improves geometric accuracy, directional consistency, invisible-region completion, render-space agreement, and physical plausibility over representative baselines.

\paragraph{Limitations.}
SAFE-Hair remains limited by the ambiguity of single-view reconstruction. Severe occlusions, unusual hairstyles, dense accessories, or extreme out-of-distribution grooming patterns can lead to inaccurate invisible-region completion. The current model is designed for static images and does not explicitly enforce temporal consistency for video. In addition, although SAFE-Hair exports simulation-oriented strands, production use may still require solver-specific tuning, material calibration, or artist editing. Future work will explore stronger completion priors, temporal modeling, and closer integration with production grooming and simulation tools.

\begin{ack}
The work is supported in part by the Early Career Scheme of the Research Grants Council (RGC) of the Hong Kong SAR under grant No. 26202321, ITF PRP/046/24FX, the Science \& Technology Cooperation Program of Shandong under grant No. SDST26EG01, SAIL Research Project, and HKUST-Zeekr Collaborative Research Fund.
\end{ack}

\newpage
{
\small
\bibliographystyle{plainnat}
\bibliography{main}
}

\newpage
\appendix
\section*{Appendix Overview}

This appendix provides additional details for SAFE-Hair and HairBench-3D.
Sec.~\ref{app:method_details} describes implementation details of SAFE-Hair, including canonical scalp-field construction, strand supervision, scalp-conditioned Rectified Flow inference, sampling steps and scalability, training objectives, geometric-validity priors, and export metadata.
Sec.~\ref{app:hairbench} presents the construction and evaluation protocol of HairBench-3D, including data sources, split construction, canonical registration, metric definitions, baseline conversion, asset-usability metrics, and robustness to head-and-camera fitting errors.
Sec.~\ref{app:additional_qualitative} provides additional qualitative results on in-the-wild portrait images.
Together, these sections are intended to make the method, benchmark, and evaluation protocol easier to reproduce.

\section{Additional Method Details}
\label{app:method_details}

\subsection{Motivation}
We make a distinction between reconstructing hair geometry and reconstructing a usable hair asset. A usable groom is not merely a visually plausible surface or collection of curves: it should contain scalp-attached roots, persistent strand or follicle identities, and a structure that can be serialized, edited, and simulated in downstream pipelines. These requirements are especially important for operations such as guide selection, child-strand interpolation, collision handling, and temporal editing, all of which are naturally defined relative to scalp-rooted strand identities. Existing single-view methods can produce compelling renderings, but their output representations do not always preserve such asset-level structure, particularly in self-occluded regions.

\subsection{Canonical scalp-field construction}
\label{app:field_construction}

Each training hairstyle is represented as a set of 3D strands and is first transformed into the coordinate frame of the fitted template head.
We use the fitted scalp surface as a shared domain for assigning raw strands to a fixed set of canonical follicles.
Let \(\mathcal{R}=\{\mathbf{u}_r\}_{r=1}^{R}\) denote canonical follicles in scalp UV space, and let \(\psi(\mathbf{u}_r)\) be the corresponding 3D scalp point.
For each ground-truth strand \(\mathbf{X}^{\star}_i\), we assign its root to the nearest canonical follicle by scalp geodesic distance:
\begin{equation}
r(i)=
\arg\min_{r}
d_{\mathrm{geo}}
\left(
\mathbf{x}^{\star}_{i,1},
\psi(\mathbf{u}_r)
\right).
\label{eq:gt_assignment}
\end{equation}
A follicle is marked as occupied if at least one strand is assigned to it.
When multiple strands are assigned to the same follicle, we keep the strand whose root is closest to the canonical follicle location. Each occupied follicle stores an arc-length-resampled strand
\(\mathbf{X}^{\star}_r=\{\mathbf{x}^{\star}_{r,t}\}_{t=1}^{T}\).
From this strand we compute the total length \(L^{\star}_r\), normalized segment-length weights \(q^{\star}_{r,t}\), and unit segment directions \(\mathbf{d}^{\star}_{r,t}\).
Specifically, \(L^{\star}_r\) is the sum of segment lengths, \(q^{\star}_{r,t}\) is each segment length divided by \(L^{\star}_r+\epsilon\), and \(\mathbf{d}^{\star}_{r,t}\) is the normalized vector from \(\mathbf{x}^{\star}_{r,t}\) to \(\mathbf{x}^{\star}_{r,t+1}\).
This decomposition matches the decoder in the main paper: length controls the global extent, segment weights distribute length along the curve, and directions encode local flow.

The per-follicle quantities are then rasterized to the scalp UV grid.
Each UV cell stores local statistics of nearby follicles, including occupancy and compressed geometry features.
A lightweight encoder maps this rasterized representation to the latent target field \(z_1\) used by the Rectified Flow model.
Empty follicles contribute occupancy labels but are ignored by geometry and descriptor losses.
Using a latent scalp field, rather than decoding all strands independently, lets neighboring follicles share context and improves completion in weakly observed regions.

\subsection{Strand-level supervision}
\label{app:strand_supervision}

The direct strand supervision term is 
\begin{equation}
\mathcal{L}_{\mathrm{strand}}
=
\lambda_{\mathrm{occ}}\mathcal{L}_{\mathrm{occ}}
+
\lambda_L\mathcal{L}_L
+
\lambda_q\mathcal{L}_q
+
\lambda_d\mathcal{L}_d
+
\lambda_{\mathrm{pos}}\mathcal{L}_{\mathrm{pos}}.
\label{eq:appendix_strand_loss}
\end{equation}
We use binary cross-entropy for occupancy, \(L_1\) losses for total length and segment-length weights, a cosine loss for directions, and an \(L_1\) loss for integrated control points.
The occupancy loss is evaluated over all canonical follicles, while geometry losses are evaluated only on occupied follicles.

Occupancy determines which follicles produce active strands and therefore affects the hairline, parting region, and scalp coverage.
At inference time, a follicle is kept if its predicted occupancy probability exceeds \(\tau_{\mathrm{occ}}\).
Length and segment-allocation terms supervise complementary aspects of the strand: the former controls coarse extent, while the latter controls how length is distributed along the curve.
The direction loss preserves local flow, and the point-level loss anchors the integrated curve in 3D space. In all experiments, the sub-loss weights are selected on the validation split and fixed for final evaluation:
\[
\lambda_{\mathrm{occ}}=1.0,\quad
\lambda_L=0.5,\quad
\lambda_q=0.2,\quad
\lambda_d=1.0,\quad
\lambda_{\mathrm{pos}}=1.0.
\]
The HairBench-3D test split is not used for loss-weight selection.
Unless otherwise specified, we keep follicles with \(o_r>\tau_{\mathrm{occ}}\), where \(\tau_{\mathrm{occ}}=0.5\).

\subsection{Scalp conditioning and Rectified Flow inference details}
\label{app:conditioning_flow_details}

We fit a template head and camera to each input portrait and use the recovered projection to lift image features into the canonical scalp-UV domain.
Dense image features are extracted with DINOv2 ViT-B/14 from a \(512\times512\) input crop.
For each scalp-UV location, the corresponding 3D scalp point is projected into the image, and the image feature map is bilinearly sampled.
Visibility is computed from the fitted head geometry and camera projection.

UV cells that are not directly visible are filled by nearest-neighbor propagation from visible scalp cells and are accompanied by a binary visibility mask.
The projected feature map, visibility mask, and positional encodings are concatenated and passed to the conditional Rectified Flow model.

The Rectified Flow network predicts \(v_\theta(z_t,t,C)\), where \(C\) is the scalp-aligned conditioning map.
At inference time, we sample \(z_0\sim\mathcal{N}(0,I)\) and solve the ODE from \(t=0\) to \(t=1\) with 20 Euler steps.
Inference therefore uses a fixed number of network evaluations and no per-subject test-time optimization.
For evaluation, we use one random sample with a fixed seed, without best-of-\(K\) selection.
Multiple samples can produce different plausible completions for unobserved side and back regions.
Over five seeds, the standard deviation is small for all metrics (Table~\ref{tab:hairbench_main}), and variations concentrate in unobserved regions.

\subsection{Sampling steps and scalability}
\label{app:steps_scalability}

Quality largely saturates at 20 Euler steps, which we use by default (Table~\ref{tab:steps}).
Increasing the follicle count \(N\) or the control points per strand \(T\) improves fidelity at roughly linear decoding cost (Table~\ref{tab:scalability}); the Rectified Flow cost is unchanged because the latent scalp field stays at \(128\times128\).

\begin{table}[h]
    \centering
    \small
    \begin{tabular}{c c c c c}
        \toprule
        Steps 
        & Time (s) $\downarrow$ 
        & ChamferPts $\downarrow$ 
        & F-score $\uparrow$ 
        & F-score (Invisible) $\uparrow$ \\
        \midrule
        5  & 1.4 & 3.47 & 79.8 & 67.4 \\
        10 & 2.4 & 3.34 & 81.0 & 69.0 \\
        20 & 4.2 & 3.28 & 81.7 & 69.8 \\
        40 & 7.8 & 3.26 & 81.9 & 70.0 \\
        \bottomrule
    \end{tabular}
    \vspace{5pt}
    \caption{\textbf{Effect of Rectified Flow sampling steps on HairBench-3D.}}
    \label{tab:steps}
\end{table}

\begin{table}[h]
    \centering
    \small
    \begin{tabular}{c c c c c c}
        \toprule
        Follicles \(N\) 
        & Points \(T\) 
        & Time (s) $\downarrow$ 
        & Memory (GB) $\downarrow$ 
        & ChamferPts $\downarrow$ 
        & F-score $\uparrow$ \\
        \midrule
        32,768 & 32 & 4.2 & 7.9  & 3.28 & 81.7 \\
        65,536 & 32 & 6.8 & 12.7 & 3.22 & 82.1 \\
        32,768 & 64 & 5.6 & 10.8 & 3.21 & 82.0 \\
        65,536 & 64 & 9.4 & 18.6 & 3.17 & 82.5 \\
        \bottomrule
    \end{tabular}
    \vspace{5pt}
    \caption{\textbf{Scalability in follicle count \(N\) and control points per strand \(T\).}}
    \label{tab:scalability}
\end{table}

\subsection{Training objectives and geometric-validity priors}
\label{app:objectives_priors}

For real portraits without 3D strand ground truth, we supervise the prediction by differentiable rendering.
The predicted strands are rendered from the estimated input camera and compared with image-derived hair cues.
The silhouette term compares the rendered hair mask with the observed or estimated hair mask, constraining the visible hair volume and outer contour.
The orientation term compares the rendered 2D strand orientation field with an image-derived orientation field over visible hair pixels, providing local flow supervision.
Since a single portrait observes only part of the hair, these image-space losses are complemented by the conditional scalp-field prior, which learns plausible completions from 3D hair data.

The geometric-validity loss is
\begin{equation}
\mathcal{L}_{\mathrm{valid}}
=
\lambda_{\mathrm{att}}\mathcal{L}_{\mathrm{att}}
+
\lambda_{\mathrm{col}}\mathcal{L}_{\mathrm{col}}
+
\lambda_{\mathrm{bend}}\mathcal{L}_{\mathrm{bend}}.
\label{eq:appendix_valid_loss}
\end{equation}
These terms reduce common asset-level failures but are not intended to replace physical simulation.
They are evaluated on decoded strands during training and require no iterative correction at test time.

The attachment prior encourages the first strand segment to leave the scalp in a plausible direction:
\begin{equation}
\mathcal{L}_{\mathrm{att}}
=
\frac{1}{|\mathcal{R}^+|}
\sum_{r\in\mathcal{R}^+}
\max
\left(
0,
\gamma_{\mathrm{att}} -
\mathbf{n}_r^\top \hat{\mathbf{d}}_{r,1}
\right),
\label{eq:attachment_loss}
\end{equation}
where \(\mathbf{n}_r\) is the outward scalp normal and \(\mathcal{R}^+\) is the set of occupied follicles.
The collision prior penalizes decoded samples that enter the fitted head mesh or lie within a small safety margin:
\begin{equation}
\mathcal{L}_{\mathrm{col}}
=
\frac{1}{|\hat{\mathcal{X}}|}
\sum_{\hat{\mathbf{x}}\in \hat{\mathcal{X}}}
\max
\left(
0,
\delta_{\mathrm{col}} - s_M(\hat{\mathbf{x}})
\right),
\label{eq:collision_loss}
\end{equation}
where \(s_M\) is the signed distance to the fitted head mesh.
The bending prior regularizes high-frequency direction changes:
\begin{equation}
\mathcal{L}_{\mathrm{bend}}
=
\frac{1}{|\mathcal{R}^+|(T-2)}
\sum_{r\in\mathcal{R}^+}
\sum_{t=1}^{T-2}
\left(
1 -
\hat{\mathbf{d}}_{r,t}^{\top}
\hat{\mathbf{d}}_{r,t+1}
\right).
\label{eq:bending_loss}
\end{equation}
We use
\[
\lambda_{\mathrm{att}}=0.1,\quad
\lambda_{\mathrm{col}}=0.5,\quad
\lambda_{\mathrm{bend}}=0.05,\quad
\gamma_{\mathrm{att}}=0.15,\quad
\delta_{\mathrm{col}}=2\,\mathrm{mm}.
\]
These values are selected on the validation split and fixed for all experiments.

\subsection{Export metadata and guide-family construction}
\label{app:export_details}

The decoder may optionally predict a low-dimensional descriptor \(\mathbf{e}_r\) for each active follicle.
This descriptor is used as export metadata rather than as the primary reconstruction target.
Depending on available supervision, it may include geometry-derived quantities such as total length, mean direction, curvature statistics, curl, or other groom-control channels.
When no descriptor targets are available, the descriptor loss is disabled.
For descriptor supervision, we use uniform descriptor weights \(w_m=1\) and global weight \(\lambda_{\mathrm{desc}}=0.01\).

When curl supervision is used, we define curl as the average change of tangent direction along the resampled strand:
\begin{equation}
c^{\star}_r
=
\frac{1}{T-2}
\sum_{t=1}^{T-2}
\arccos
\left(
\mathrm{clip}
\left(
{\mathbf{d}^{\star}_{r,t}}^{\top}
\mathbf{d}^{\star}_{r,t+1},
-1,
1
\right)
\right).
\label{eq:curl_descriptor}
\end{equation}
Descriptor supervision is applied only to occupied follicles and only to dimensions with available targets:
\begin{equation}
\mathcal{L}_{\mathrm{desc}}
=
\frac{1}{|\mathcal{R}^{+}|}
\sum_{r \in \mathcal{R}^{+}}
\sum_{m \in \mathcal{M}_r}
w_m
\left|
e_{r,m}
-
e^{\star}_{r,m}
\right|.
\label{eq:appendix_desc_loss}
\end{equation}

\begin{figure}[h]
    \centering
    \begin{subfigure}{0.26\linewidth}
        \centering
        \includegraphics[width=\linewidth]{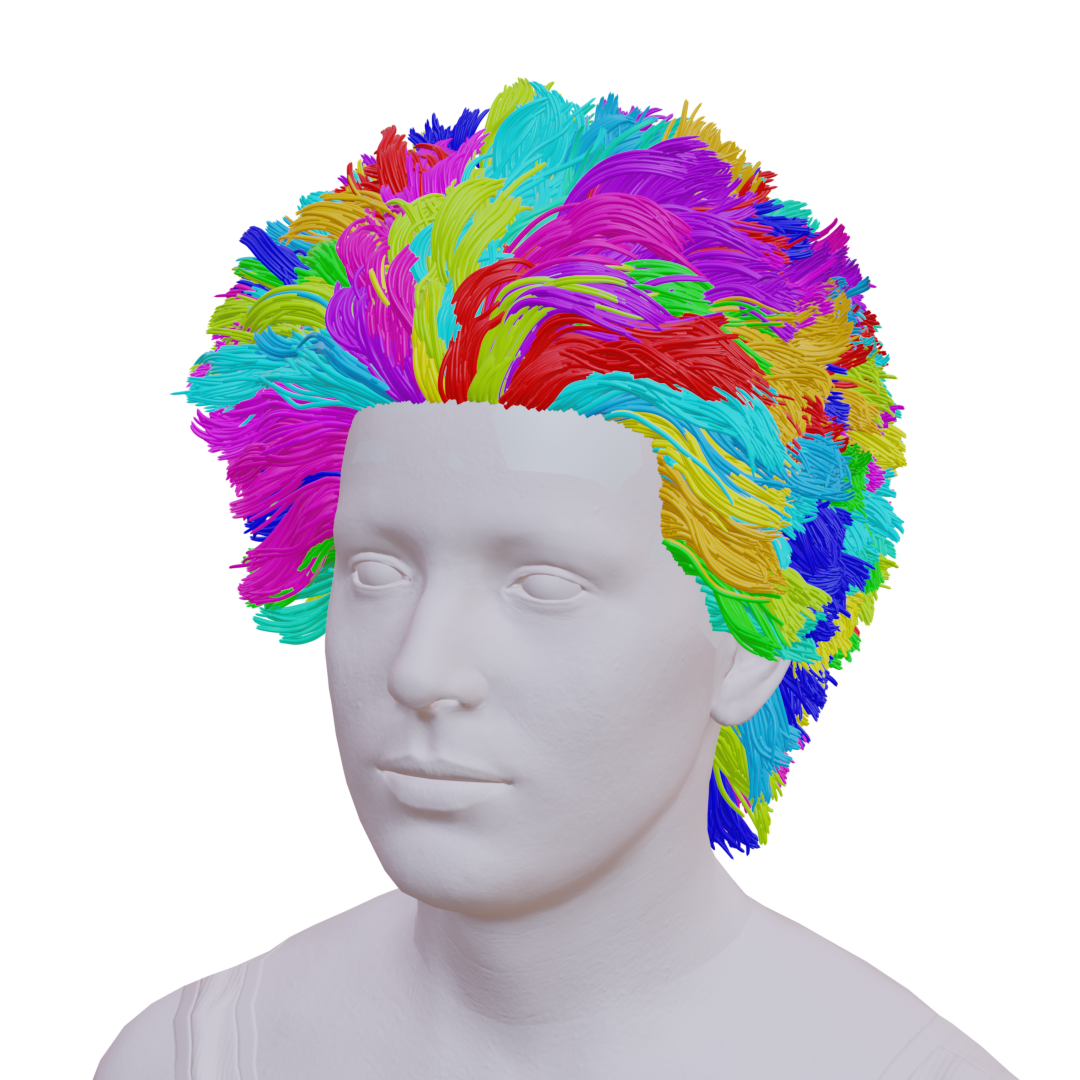}
        \caption{Cluster}
        \label{fig:guide_cluster}
    \end{subfigure}
    \hspace{4mm}
    \begin{subfigure}{0.26\linewidth}
        \centering
        \includegraphics[width=\linewidth]{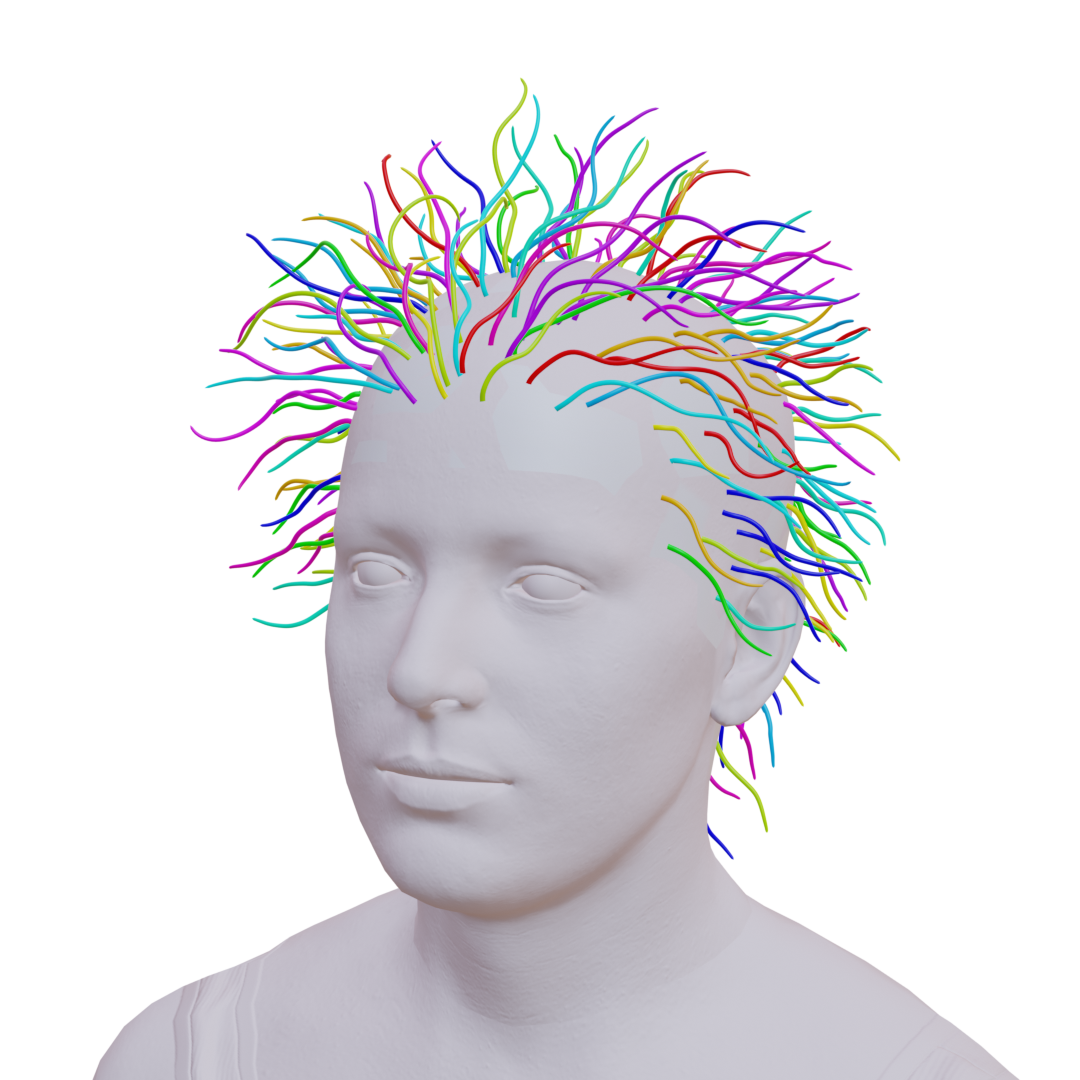}
        \caption{Guide strands}
        \label{fig:guide_strands}
    \end{subfigure}
    \hspace{4mm}
    \begin{subfigure}{0.26\linewidth}
        \centering
        \includegraphics[width=\linewidth]{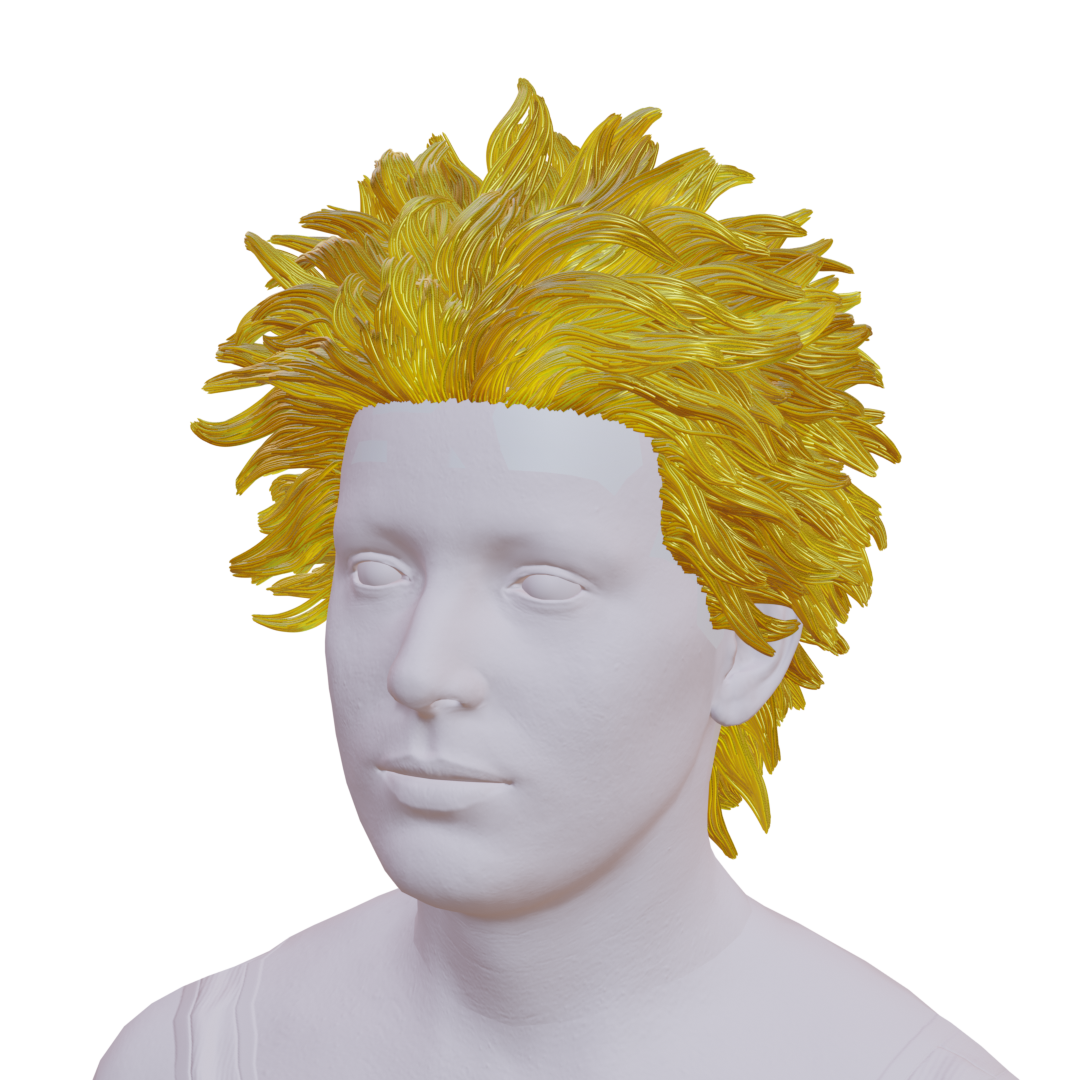}
        \caption{Refinement}
        \label{fig:guide_refinement}
    \end{subfigure}
    \caption{\textbf{Exportable groom structure.}
    SAFE-Hair groups active follicles into guide families in canonical scalp UV space.
    Each child strand stores a persistent follicle ID, guide ID, control points, and optional physical or groom-control attributes.}
    \label{fig:guide_family_structure}
\end{figure}

Guide-family construction is performed after strand reconstruction and does not affect predicted geometry.
We select guide strands from active follicles using farthest-point sampling in scalp UV space.
Optionally, the sampling score can be weighted by strand length or curvature so that visually important strands are more likely to become guides.
Each remaining strand is assigned to the nearest guide using the joint scalp-position and strand-feature distance from the main paper.
The exported asset stores control points, follicle ID, scalp UV coordinate, optional descriptor, and optional guide ID, which supports curve-based groom serialization and downstream editing.

\section{HairBench-3D Benchmark Details}
\label{app:hairbench}

\subsection{Data Sources and Licenses}
\label{app:hairbench_sources}

HairBench-3D is constructed from strand-level hairstyle assets collected from public or research-available sources, including Hair20K, USC-HairSalon, and Yuksel Hair Meshes~\citep{usc_hairsalon,zhou2024hair20k,yuksel2009hairmeshes}.
We use only assets that can be converted to explicit 3D strands with root positions, ordered control points, and strand tangents.
Assets with missing geometry, ambiguous scale, corrupted strand order, or severe head misalignment are excluded.

For each source, we record the original asset identifier, source dataset, license or usage terms, and processed split assignment.
The benchmark release contains only assets or derived annotations permitted by the source terms.
When raw asset redistribution is not permitted, we provide preprocessing scripts, split files, and instructions for obtaining the original data from the official source. Because different sources may contain shared or near-identical hairstyles, we perform duplicate and near-duplicate detection before assigning train, validation, or test membership.
The official test split is frozen and is not modified after model development begins.

\subsection{Split Construction and Leakage Prevention}
\label{app:hairbench_split}

We construct the split at the asset level rather than the rendered-image level.
All rendered views, augmentations, and processed variants of the same hairstyle inherit the same split assignment.
This prevents different views of one hairstyle from appearing in both training and test sets.

We compute asset descriptors for duplicate detection, including scalp-root distributions, strand-length histograms, root-direction histograms, rendered silhouettes from canonical views, and global geometric statistics.
Potential near-duplicates are identified using these descriptors and are manually inspected or conservatively assigned to the same split.

The final HairBench-3D test split contains 1,551 hairstyles.
No test asset, rendered input, augmented variant, near-duplicate, or processed derivative is included in the SAFE-Hair training or validation pool.
The test split is not used for hyperparameter selection, loss-weight tuning, filtering-threshold selection, early stopping, model selection, or qualitative example selection.

\begin{table}[h]
    \centering
    \small
    \begin{tabular}{l c}
        \toprule
        Usage of HairBench-3D test split & Allowed? \\
        \toprule
        Training data & No \\
        Validation or hyperparameter tuning & No \\
        Loss-weight selection & No \\
        Filtering-threshold selection & No \\
        Qualitative example selection before final evaluation & No \\
        Final quantitative evaluation & Yes \\
        Final qualitative visualization of evaluated samples & Yes \\
        \toprule
    \end{tabular}
    \vspace{5pt}
    \caption{Usage policy for the frozen HairBench-3D test split.}
    \label{tab:test_usage_policy}
\end{table}

\subsection{Canonical Registration and Scalp-UV Parameterization}
\label{app:hairbench_registration}

All benchmark assets are converted to a shared canonical coordinate system using a template head with a predefined scalp mask and scalp-UV parameterization.
For each hairstyle, we estimate a rigid alignment to the template head using the associated head mesh or verified landmark correspondences.
Each strand root is then projected onto the closest point on the template scalp.

A strand is considered valid if its projected root lies inside the scalp mask and its root-to-scalp distance is below a fixed threshold.
Invalid strands are discarded.
Each valid strand is assigned a canonical follicle index by nearest-neighbor lookup on the scalp surface and resampled to \(T\) ordered control points by arc length.
We store the resampled points, unit tangents, total length, root location, follicle index, and visibility labels. This canonicalization ensures that all methods are evaluated under the same scale, orientation, and head geometry.
It also enables root-aware, direction-aware, and invisible-region metrics across methods with different native representations.

\subsection{Metric Definitions}
\label{app:metrics}

Let \(P=\{(p_i, d_i)\}\) denote predicted strand samples and \(G=\{(g_j, t_j)\}\) denote ground-truth samples.
All points are expressed in the canonical coordinate system and distances are measured in millimeters.

\paragraph{ChamferPts.}
ChamferPts measures symmetric point-level distance:
\begin{equation}
\mathrm{ChamferPts}(P,G)
=
\frac{1}{|P|}\sum_{p_i\in P}\min_{g_j\in G}\|p_i-g_j\|_2
+
\frac{1}{|G|}\sum_{g_j\in G}\min_{p_i\in P}\|g_j-p_i\|_2 .
\label{eq:app_chamfer}
\end{equation}
We uniformly sample strand points by arc length before computing the metric.

\paragraph{\(\mathrm{F}\text{-}\mathrm{score}@4\,\mathrm{mm}/40^\circ\).}
A predicted sample is matched if there exists a ground-truth sample satisfying
\begin{equation}
\|p_i-g_j\|_2 < 4\,\mathrm{mm}
\quad \text{and} \quad
\arccos(|d_i^\top t_j|) < 40^\circ .
\label{eq:app_fscore_match}
\end{equation}
Precision and recall are computed from matched predicted and ground-truth samples, and the F-score is
\begin{equation}
F =
\frac{2 \cdot \mathrm{precision}\cdot \mathrm{recall}}
{\mathrm{precision}+\mathrm{recall}+\epsilon}.
\label{eq:app_fscore}
\end{equation}

\paragraph{Invisible-region F-score.}
Invisible-region F-score uses the same distance and tangent thresholds as above but restricts ground-truth samples to those invisible from the input view.
Visibility labels are computed using z-buffer visibility on rendered ground-truth strand samples.
This metric evaluates completion in self-occluded side and back regions.

\paragraph{Silhouette IoU.}
We render predicted and ground-truth hairstyles from held-out views using the same camera and head geometry.
Silhouette IoU is
\begin{equation}
\mathrm{IoU}
=
\frac{|M_{\mathrm{pred}}\cap M_{\mathrm{gt}}|}
{|M_{\mathrm{pred}}\cup M_{\mathrm{gt}}|}.
\label{eq:app_iou}
\end{equation}

\paragraph{Penetration Ratio.}
Penetration Ratio measures the fraction of predicted samples inside the head or body proxy:
\begin{equation}
\mathrm{Penetration}
=
\frac{1}{|P|}
\sum_{p_i\in P}
\mathbf{1}
\left[
\mathrm{SDF}_{\mathrm{head/body}}(p_i)<0
\right].
\label{eq:app_penetration}
\end{equation}
Lower values indicate fewer interpenetration artifacts.

\subsection{Baseline Output Conversion}
\label{app:baseline_protocol}

HairBench-3D supports methods with different output representations.
For fair evaluation, each output is converted to the representation required by each metric using fixed, method-agnostic rules.

For strand-based methods, we resample predicted strands by arc length and compute tangents from adjacent control points.
For mesh-based methods, we uniformly sample surface points and estimate local directions from available strand directions, anisotropic texture orientation, or local principal directions.
For implicit or volumetric methods, we extract the hair surface or occupancy boundary using the authors' recommended procedure and sample points from the extracted geometry.
For Gaussian- or particle-based representations, we use Gaussian centers or particle locations as point samples and estimate directions if available.

For direction-aware metrics, methods without reliable tangents are evaluated using their best available local orientation estimate.
If no reliable orientation is available, we report distance-only metrics in the supplement and mark direction-aware metrics as not directly applicable.
In the main comparison, we include only baselines whose outputs can be converted to the required metric inputs without using ground-truth 3D hair.
All conversion code and sampling densities are fixed across methods.

Because prior methods are trained on different private, synthetic, or partially released collections, HairBench-3D should be interpreted as a fixed held-out system-level evaluation rather than a claim of identical training data.
When trainable code is available, we additionally evaluate retrained or fine-tuned variants under the same training split; otherwise, we use official released weights and document this choice. HairOrbit~\cite{jin2026hairorbit} and TANGLED \cite{long2025tangled} have released their code for fair comparison.

\subsection{Additional asset-usability metrics}
\label{app:asset_usability_metrics}

HairBench-3D also supports asset-usability metrics that evaluate whether a reconstruction behaves as a scalp-rooted strand groom. Results are reported in Table~\ref{tab:asset_usability}.
ExportSucc.\ is the percentage of predictions that can be serialized as valid strand-groom assets, and ``--'' marks metrics that a method's output cannot provide.
For DiffLocks and Im2Haircut, follicle stability is approximated by matching each predicted root to its nearest canonical follicle.
These methods can also be exported after post-processing, whereas SAFE-Hair provides scalp attachment and follicle identities natively.

\paragraph{Root-to-scalp distance.}
For each active predicted strand, we compute the distance from its root to the canonical scalp surface:
\begin{equation}
d_{\mathrm{root}}
=
\frac{1}{|\hat{\mathcal{R}}^+|}
\sum_{r\in\hat{\mathcal{R}}^+}
\mathrm{dist}
\left(
\hat{\mathbf{x}}_{r,1},
M_{\mathrm{scalp}}
\right).
\label{eq:root_distance}
\end{equation}

\paragraph{Valid-root ratio.}
We report the fraction of active strands whose roots lie within the scalp mask and within a distance threshold \(\tau_{\mathrm{root}}\):
\begin{equation}
\mathrm{ValidRoot}
=
\frac{1}{|\hat{\mathcal{R}}^+|}
\sum_{r\in\hat{\mathcal{R}}^+}
\mathbf{1}
\left[
\hat{\mathbf{x}}_{r,1}\in M_{\mathrm{scalp}}
\;\wedge\;
\mathrm{dist}(\hat{\mathbf{x}}_{r,1},M_{\mathrm{scalp}})<\tau_{\mathrm{root}}
\right].
\label{eq:valid_root_ratio}
\end{equation}
We use \(\tau_{\mathrm{root}}=2\,\mathrm{mm}\) in all experiments.

\paragraph{Follicle stability under input perturbations.}
To measure follicle-identity stability, we evaluate the same image under small crop, color, or segmentation perturbations.
Given active follicle sets \(\hat{\mathcal{R}}^+_a\) and \(\hat{\mathcal{R}}^+_b\), we compute
\begin{equation}
\mathrm{FollicleIoU}
=
\frac{
|\hat{\mathcal{R}}^+_a \cap \hat{\mathcal{R}}^+_b|
}{
|\hat{\mathcal{R}}^+_a \cup \hat{\mathcal{R}}^+_b|
}.
\label{eq:follicle_iou}
\end{equation}

\paragraph{Guide-family compactness.}
For exported guide families, we measure the average child-to-guide distance in the joint scalp-position and strand-feature space:
\begin{equation}
\mathrm{GuideCompact}
=
\frac{1}{|\hat{\mathcal{R}}^+|}
\sum_{r\in\hat{\mathcal{R}}^+}
\left(
\alpha \|\mathbf{u}_r-\mathbf{u}_{g(r)}\|_2^2
+
\beta \|\mathbf{f}_r-\mathbf{f}_{g(r)}\|_2^2
\right).
\label{eq:guide_compact}
\end{equation}

\subsection{Robustness to head-and-camera fitting errors}
\label{app:robustness_details}

We perturb the estimated camera by yaw, pitch, and roll offsets sampled from \(\mathcal{U}(-5^\circ,5^\circ)\), and the estimated head shape by Gaussian noise (\(0.5\times\) the training-set standard deviation) on the first ten PCA coefficients; ``Combined'' applies both.
The same perturbations are applied to DiffLocks and Im2Haircut, and all methods are evaluated against the unperturbed ground truth, where RootDist-to-GT measures root distance to the ground-truth scalp.
SAFE-Hair degrades the least under every perturbation, most clearly in the root metrics (Tables~\ref{tab:perturb_abs} and~\ref{tab:perturb}).

\begin{table}[h]
    \centering
    \small
    \begin{tabular}{l c c c c}
        \toprule
        Perturbation 
        & ChamferPts $\downarrow$ 
        & F-score $\uparrow$ 
        & RootDist-to-GT $\downarrow$ 
        & ValidRoot $\uparrow$ \\
        \midrule
        None 
        & 3.28 & 81.7 & 0.20 & 99.8 \\
        Camera rotation ($\pm5^\circ$) 
        & 3.33 & 81.2 & 0.22 & 99.6 \\
        Head-shape deviation 
        & 3.53 & 79.5 & 0.96 & 97.8 \\
        Combined 
        & 3.62 & 78.9 & 1.15 & 96.7 \\
        \bottomrule
    \end{tabular}
    \vspace{5pt}
    \caption{\textbf{SAFE-Hair under head-and-camera fitting errors.}}
    \label{tab:perturb_abs}
\end{table}

\begin{table}[h]
    \centering
    \small
    \setlength{\tabcolsep}{4.0pt}
    \resizebox{\columnwidth}{!}{
    \begin{tabular}{l cccc cccc cccc}
        \toprule
        & \multicolumn{4}{c}{Camera rotation} & \multicolumn{4}{c}{Head-shape deviation} & \multicolumn{4}{c}{Combined} \\
        \cmidrule(lr){2-5}\cmidrule(lr){6-9}\cmidrule(lr){10-13}
        Method & $\Delta$CD & $\Delta$F & $\Delta$Root & $\Delta$Valid & $\Delta$CD & $\Delta$F & $\Delta$Root & $\Delta$Valid & $\Delta$CD & $\Delta$F & $\Delta$Root & $\Delta$Valid \\
        \midrule
        DiffLocks~\citep{rosu2025difflocks} & 2.2\% & 0.8 & 0.08 & 0.8 & 9.8\% & 3.1 & 1.25 & 4.8 & 14.2\% & 4.1 & 1.52 & 6.1 \\
        Im2Haircut~\citep{sklyarova2025im2haircut} & 2.0\% & 0.7 & 0.05 & 0.5 & 9.4\% & 2.8 & 1.02 & 3.8 & 12.6\% & 3.7 & 1.24 & 5.0 \\
        Ours & \textbf{1.5\%} & \textbf{0.5} & \textbf{0.02} & \textbf{0.2} & \textbf{7.6\%} & \textbf{2.2} & \textbf{0.76} & \textbf{2.0} & \textbf{10.4\%} & \textbf{2.8} & \textbf{0.95} & \textbf{3.1} \\
        \bottomrule
    \end{tabular}
    }
    \vspace{5pt}
    \caption{\textbf{Robustness to head-and-camera fitting errors.}
    Degradation from each method's clean setting (lower is better); $\Delta$F and $\Delta$Valid in points, $\Delta$Root in mm.}
    \label{tab:perturb}
\end{table}

\section{Additional Qualitative Results}
\label{app:additional_qualitative}

\begin{figure}[ht]
    \centering
    \includegraphics[width=0.4\linewidth]{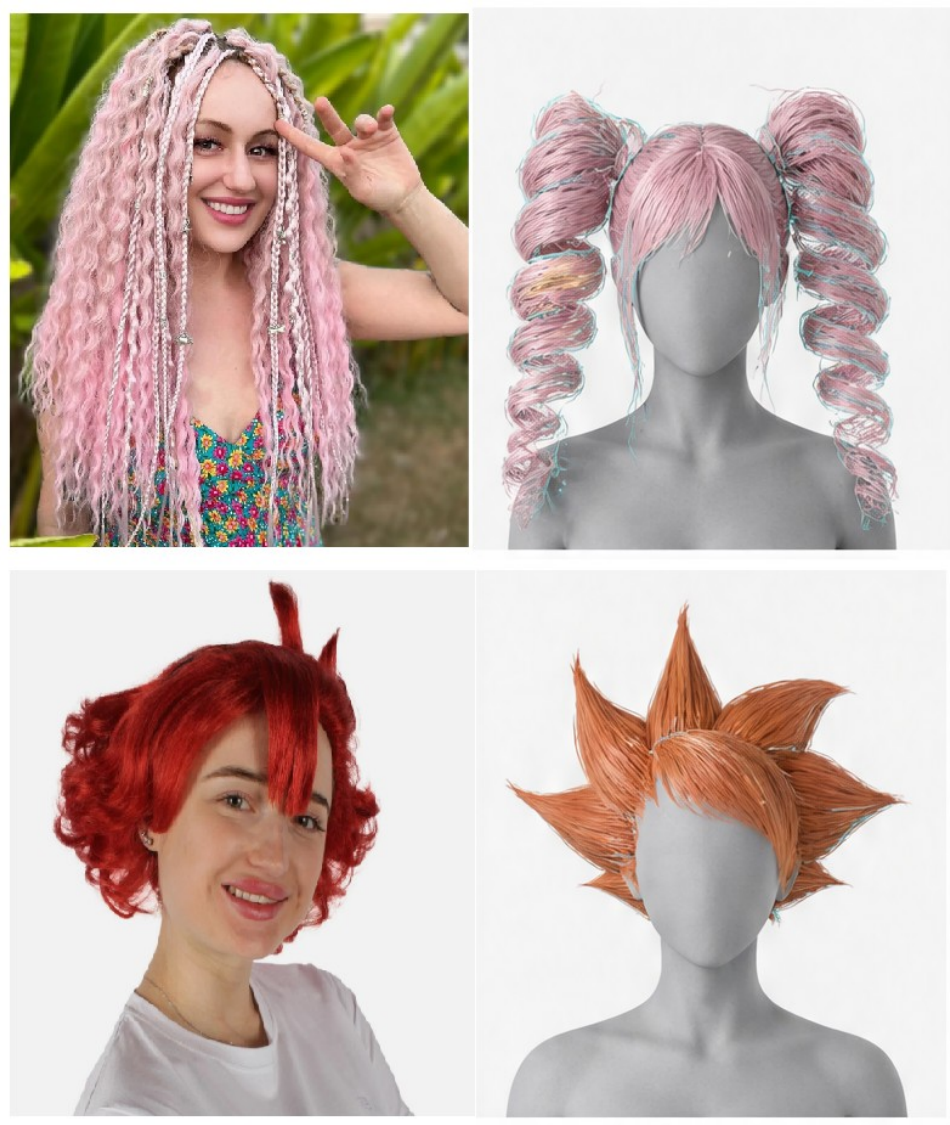}
    \caption{\textbf{Limitation.} Extreme hairstyles.}
    \label{fig:limitation}
\end{figure}

\begin{figure}[t]
    \centering
    \includegraphics[width=\textwidth]{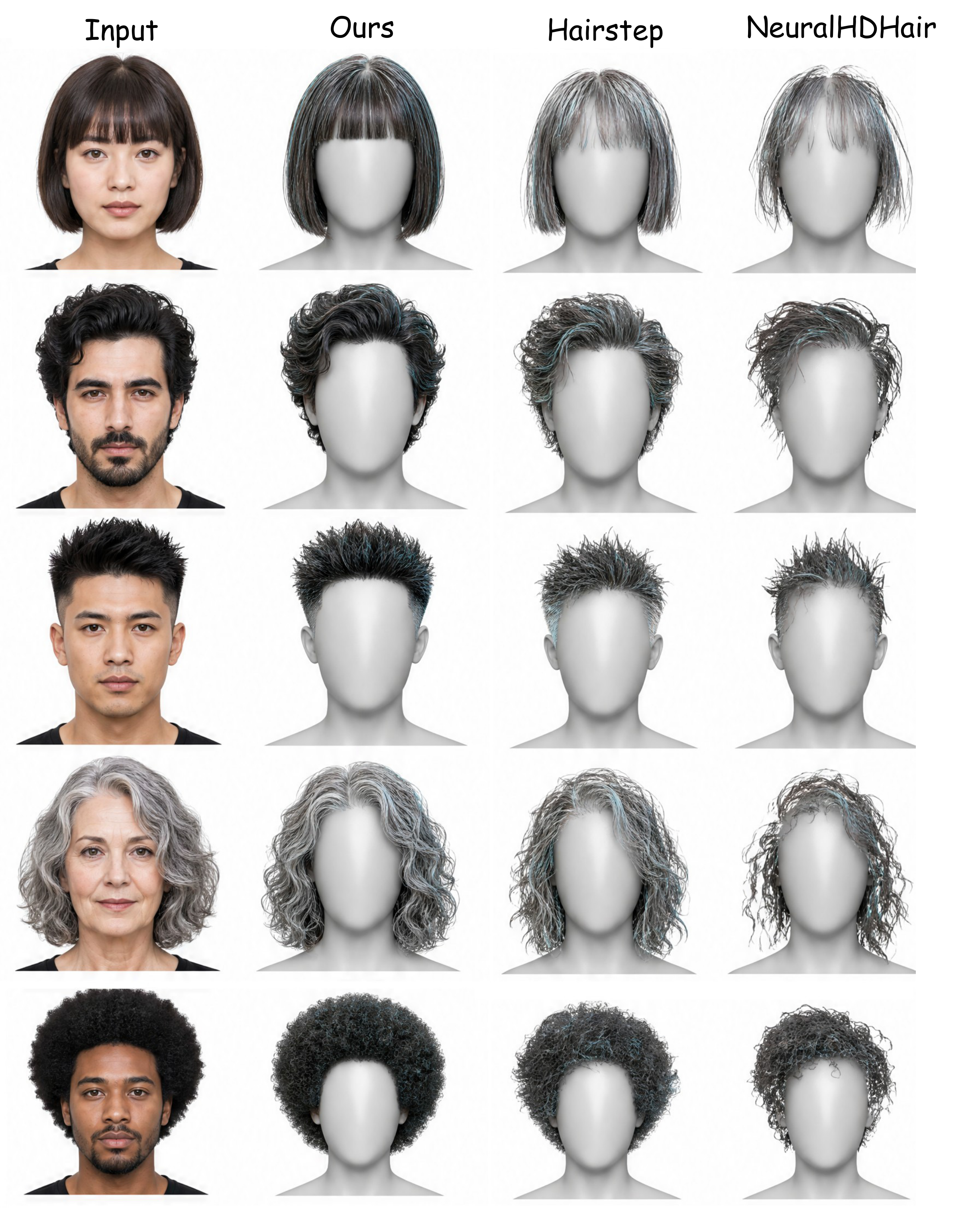}
    \caption{
    Additional qualitative comparison on in-the-wild portrait images.
    From left to right: input image, SAFE-Hair, Hairstep, and NeuralHDHair.
    SAFE-Hair produces more coherent and complete strand reconstructions across diverse hairstyles, including short, straight, wavy, spiky, curly, and afro hair.
    Compared with baselines, our method better preserves the global hair silhouette and local strand flow, while reducing floating strands, sparse regions, and unstable scalp attachment.
    }
    \label{fig:appendix_in_the_wild_comparison}
\end{figure}

\begin{figure}[h]
    \centering
    \includegraphics[width=\textwidth]{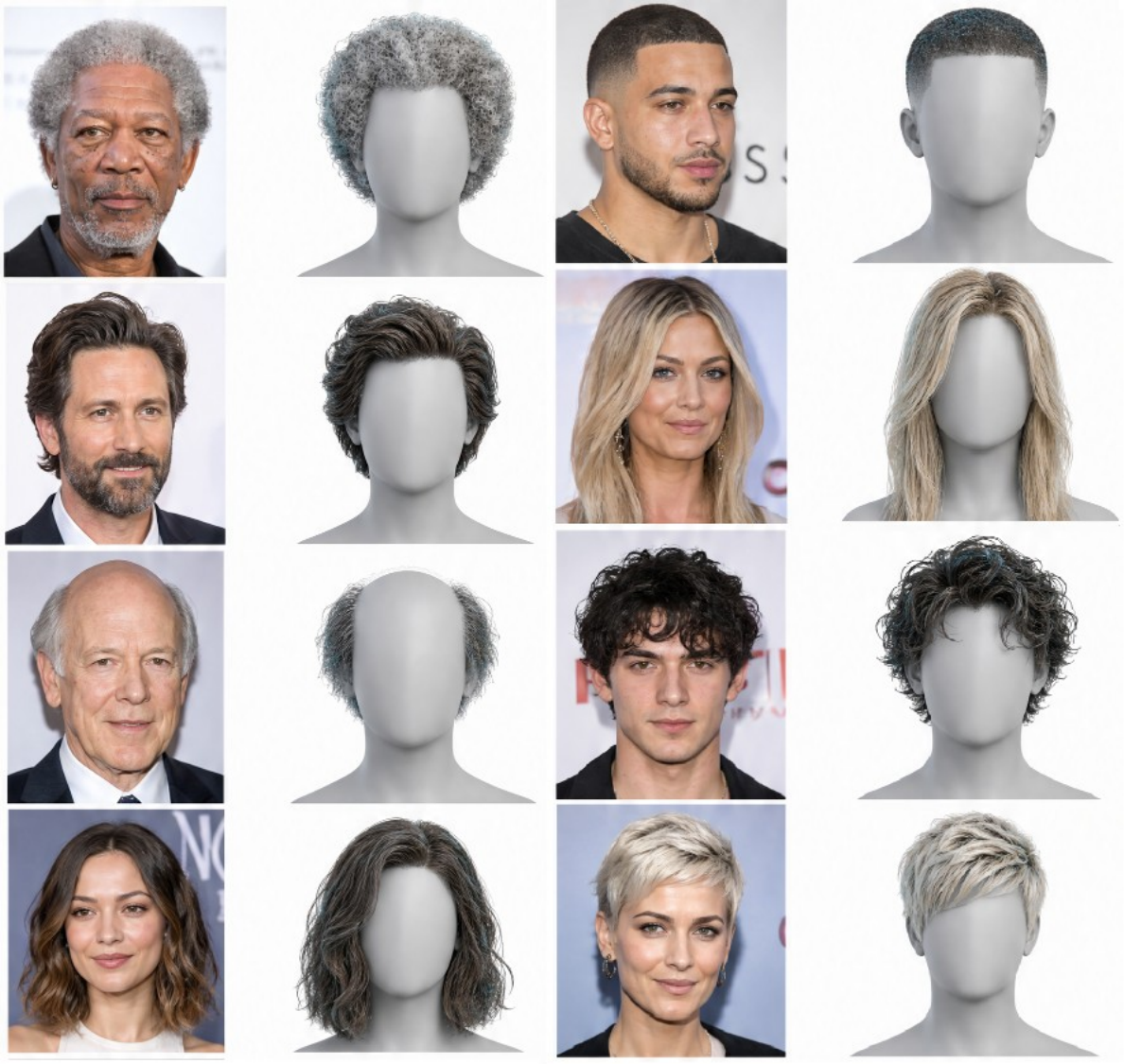}
    \caption{
    \textbf{Additional qualitative results on in-the-wild portrait images.}
    For each pair, the left image shows the input portrait and the right image shows the reconstructed 3D hair rendered on a neutral template head.
    These examples show the diversity of single-image reconstructions produced by SAFE-Hair.
    }
    \label{fig:appendix_ood_portraits}
\end{figure}

The qualitative examples show SAFE-Hair reconstructions from in-the-wild portrait images with diverse hairstyles, identities, lighting conditions, and backgrounds.
For each example, we visualize the input portrait and the reconstructed 3D hair rendered on a neutral template head. Results are shown in Figures~\ref{fig:appendix_in_the_wild_comparison} and~\ref{fig:appendix_ood_portraits}.
Figure~\ref{fig:limitation} shows extreme hairstyles that remain challenging for single-view reconstruction, as discussed in the limitations of Sec.~\ref{sec:conclusion}.

\end{document}